\documentclass[sigconf]{acmart}
\AtBeginDocument{%
  }

\setcopyright{acmlicensed}
\copyrightyear{2018}
\acmYear{2018}
\acmDOI{XXXXXXX.XXXXXXX}
\acmConference[Conference acronym 'XX]{Make sure to enter the correct
  conference title from your rights confirmation emai}{June 03--05,
  2018}{Woodstock, NY}
\acmISBN{978-1-4503-XXXX-X/18/06}

\usepackage{subcaption}
\usepackage{booktabs}
\usepackage{multirow}
\usepackage{makecell}
\usepackage{todonotes}
\usepackage{paralist}
\begin{document}

\title{Evaluation for Regressive Analyses on Evolving Data Streams}

\author{Yibin Sun}
\orcid{0000-0002-8325-1889}
\authornotemark[1]
\affiliation{%
  \institution{University of Waikato}
  \city{Hamilton}
  \state{Waikato}
  \country{New Zealand}
}
\email{yibin.spencer.sun@gmail.com}

\author{Heitor Murilo Gomes}
\orcid{0000-0002-5276-637X}
\affiliation{%
  \institution{Victoria University of Wellington}
  \city{Wellington}
  \country{New Zealand}}
\email{heitor.gomes@vuw.ac.nz}

\author{Bernhard Pfahringer}
\orcid{0000-0002-3732-5787}
\affiliation{%
  \institution{University of Waikato}
  \city{Hamilton}
  \country{New Zealand}}
\email{bernhard@waikato.ac.nz}

\author{Albert Bifet}
\orcid{0000-0002-8339-7773}
\affiliation{%
  \institution{University of Waikato}
  \city{Hamilton}
  \country{New Zealand}}
\email{abifet@waikato.ac.nz}

\renewcommand{\shortauthors}{Sun et al.}

\begin{abstract}
This paper explores the challenges of regression analysis in evolving data streams, an area that remains relatively underexplored compared to classification. We propose a standardized evaluation process for regression and prediction interval tasks in streaming contexts. Additionally, we introduce an innovative drift simulation strategy capable of synthesizing various drift types, including the less-studied incremental drift. Comprehensive experiments with state-of-the-art methods, conducted under the proposed process, validate the effectiveness and robustness of our approach.
\end{abstract}



\begin{teaserfigure}
\centering
  \includegraphics[width=.9\textwidth]{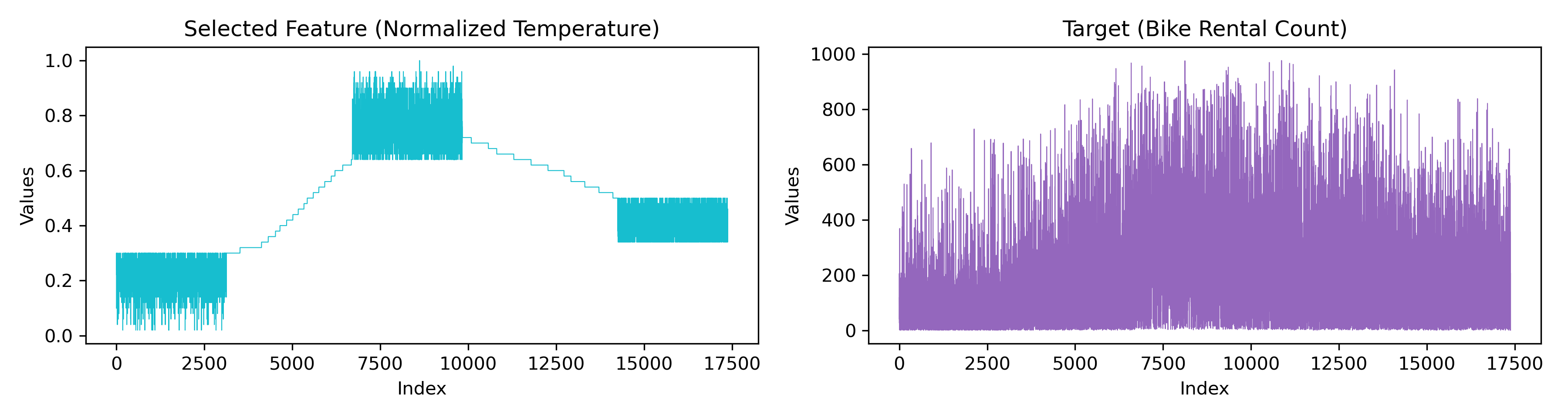}
  \caption{Simple Showcase of Feature and Target Values with Simulated Incremental Drifts}
  \Description{A demonstration of the simulation of incremental drifts with the bike dataset that includes one feature column and one target column.}
  \label{fig:teaser}
\end{teaserfigure}

\received{20 February 2007}
\received[revised]{12 March 2009}
\received[accepted]{5 June 2009}

\maketitle

\section{Introduction}
Machine learning on streaming data has garnered significant interest due to its applicability in dynamic and evolving environments~\cite{ref_stream1, ref_stream2}. However, while extensive research has focused on classification tasks in stream learning, regression tasks remain underexplored~\cite{ref_less}, with one of the reasons being a lack of dedicated data resources~\cite{ref_adapi}. The challenges are further compounded by the difficulty in defining and identifying drift in the real-world data sequences~\cite{ref_identify}. A good example for real-world applications of data stream regression is real-time energy pricing adjustments based on evolving market conditions~\cite{ref_nzep}. 
%Additionally, Prediction Intervals (PI) for streaming data have demonstrated significant effectiveness in quantifying regression uncertainty~\cite{ref_nzep}.
In addition to a point prediction, regression also generally needs uncertainty quantification, which can be provided by Prediction Intervals (PI) for streaming data~\cite{ref_nzep}.

In this work, we aim to address these gaps by making the following key deliverables:
\begin{inparaenum}
\item Standardized procedure and metrics for evaluating streaming regression algorithms,
\item Methodologies for simulating concept drifts, especially incremental drifts,
\item Empirical analysis using state-of-the-art streaming regression and prediction interval techniques.
\end{inparaenum}

All code, datasets, and scripts are publicly available on GitHub\footnote{\url{https://github.com/YibinSun/KDD25-DriftSimulation/}}, ensuring full transparency and reproducibility. This paper aims to establish a foundation for rigorous regression research in data streams while encouraging the adoption of open-source practices.

\section{Background}

This section provides basic introduction to the stream learning concepts relevant to this work.

\subsection{Data Stream}\label{sec:ds}
Data stream refers to continuous, real-time flow of data that requires dynamic processing and analysis~\cite{ref_moa}.
Typical sources of data streams include sensors, logs, online activities, and etc~\cite{ref_app}. 
Streaming data increasingly grows its significance as the world digitalizing~\cite{ref_trend}.

A data stream usually consist of a sequence of examples $DS = \{X_1, X_2, X_3, ..., X_t, ... \}$ that can be mapped to a target sequence $\mathcal{Y} \in \{ \mathcal{Y}_1, \mathcal{Y}_2, \mathcal{Y}_3, ..., \mathcal{Y}_t, ... \}$, where the subscript $t$ denotes the observation moment (time step). 
Traditionally, the $X$ is a $d$-dimensional vector and $\mathcal{Y}$ represents the ground truth at the moment $t$. In the more explored classification scenario, $\mathcal{Y}$ is usually one of the $n$ possible labels (classes), i.e., $\mathcal{Y} \in \{ C_1, C_2, ..., C_n\}$. However, in this work we focus on the regression tasks, where $\mathcal{Y}$ is a continuous value, i.e., $\mathcal{Y} \in \mathbb{R}$. 

It is worth mentioning that the label $\mathcal{Y}$ can be not available all the time. Consequently, stream learning can be categorized into several sets~\cite{ref_semi}: 
\begin{itemize}
    \item Supervised Learning: where $\mathcal{Y}$ is always immediately available after the observation moment;
    \item Unsupervised Learning: where $\mathcal{Y}$ is never available;
    \item Semi-supervised Learning: where $\mathcal{Y}$ is partially available; and
    \item Delayed Learning: where $\mathcal{Y}$ can be available at any moment posterior to the observation.
\end{itemize}
This work has a sole concentration on supervised learning. Hence, the regression $DS$ under full supervision can be represented as $DS_s = \{(X_1, \mathcal{Y}_1), (X_2, \mathcal{Y}_2), (X_3, \mathcal{Y}_3), ..., (X_t, \mathcal{Y}_t), ...\}$.

Conventional machine learning approaches always encounter issues when applying to streaming data~\cite{ref_moa}. For instance, due to the potentially infinite amount of data points, data streams cannot be stored in the machine's memory~\cite{ref_samknn}. Moreover, the high arrival velocity of the streaming data strictly restricts the processing time on each instance~\cite{ref_highspeed}. The temporal order also raises the crucial phenomenon of concept drift when the underlying distribution of the data shifts over time~\cite{ref_pitfalls}.
These constraints require the streaming algorithms to be efficient enough, can only access to an instance once (or a small amount of times), and be able to detect and adapt to the new distributions in the occurrence of concept drifts.

\subsection{Concept Drift}\label{sec:cd}
Machine learning typically assumes that the data points are independent and identically distributed (i.i.d.). However, this assumption is frequently violated in stream learning, especially with the drifting concepts.
Stream learning further assumes that the data within the a single concept should be i.i.d., yet instances from different concepts violate the i.i.d. assumption~\cite{ref_recurrent, ref_kddtutorial}.

As a consequence, concept drifts can be categorized into different types according to the behaviours when changes occur~\cite{ref_recurrent, ref_cdsurvey}.

 Abrupt (Sudden) drift refers to the sudden change from a concept to another one. An abrupt can be expressed as:
$$DS_a = \{ X^a_1, X^a_2, X^a_3, ..., X^a_t, X^b_{t+1}, X^b_{t+2}, X^b_{t+3},... \}$$ 
\noindent where $a$ and $b$ denote different concepts. At moment $t$, the data suddenly switch its source from concept $a$ to $b$.

Gradual drift refers to a drift between two concepts with a fluctuation period, which can be formulated as: 
\begin{align*}
DS_g = &\ \{ X^a_1, X^a_2, X^a_3, ..., X^a_t, X^b_{t+1}, X^a_{t+2}, \\
       &\ X^b_{t+3}, X^b_{t+4}, X^a_{t+5},..., X^b_{t'}, X^b_{t'+1}, X^b_{t'+2}, ...\}
\end{align*}
% $$DS_g = \{ X^a_1, X^a_2, X^a_3, ..., X^a_t, X^b_{t+1}, X^a_{t+2}, X^b_{t+3}, X^b_{t+4}, X^a_{t+5},..., X^b_{t'}, X^b_{t'+1}, X^b_{t'+2}, ...\}$$ 
\noindent During moment $t$ to $t+5$, a mixed data from both concept $a$ and $b$ is presented.
     
Incremental drift refers to a gradual and continuous transition between two concepts over time. This can be formulated as: $$DS_i = \{ X^a_1, X^a_2, ..., X^a_t, X^{a \rightarrow b}_{t+1}, X^{a \rightarrow b}_{t+2}, ..., \\X^{a \rightarrow b}_{t'}, X^b_{t'+1}, X^b_{t'+2}, ... \}$$
\noindent where \({a \rightarrow b}\) represents a progressive transfer from concept \(a\) to \(b\). 
    \begin{itemize}
        \item Before \(t+1\), the data is dominated by concept \(a\).  
        \item At \(t+1, t+2, ...\), the concept \(a\) is shifting to concept \(b\). 
        \item By \(t'\), the data is dominated entirely by concept \(b\).
    \end{itemize}

\begin{figure}[!ht]
    \centering
    \begin{subfigure}[b]{0.15\textwidth}
            \centering
        \includegraphics[width=\textwidth]{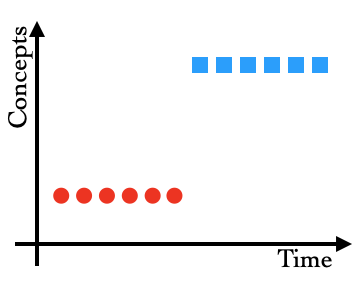} 
        \caption{Abrupt Drift}
    \end{subfigure}
    \begin{subfigure}[b]{0.15\textwidth}
        \centering
        \includegraphics[width=\textwidth]{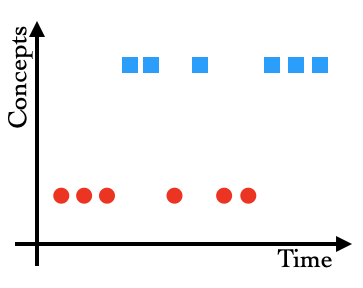}
        \caption{Gradual Drift}
    \end{subfigure}
    \begin{subfigure}[b]{0.15\textwidth}
        \centering
        \includegraphics[width=\textwidth]{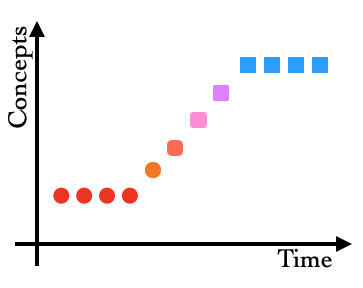}
        \caption{Incremental Drift}
    \end{subfigure}
    \caption{Illustration of Different Concept Drift Rates}
    \label{fig:driftrate}
\end{figure}

Figure~\ref{fig:driftrate} provides a visualization for the types of drift.

Concept drifts can also be regarded as different types based on the ``reach of the changing''. A drifting progress can affect feature space, target space, or both. A more detailed explanation of this can be found in~\cite{ref_nuwan}.

Recurrent drift is a more high-level and complex type of drift. It refers to the phenomenon that a particular distribution could reoccur a certain period~\cite{ref_recurrent}, where the transition between different distributions can obey any above mentioned drift type.

\subsection{Real-World Datasets}\label{sec:real-data}
The data stream community suffers from the lack of suitable real-world datasets. As criticized by Souza et al. in~\cite{ref_insect}, there are several issues challenging the stream learning with real-world datasets, including difficulty in defining drifts, bias, data stream as an afterthought, etc. 
These issues become more severe when dealing with regression problems.
The data stream regression researchers are still utilizing old and ``not-dedicated-to-stream'' datasets for evaluating new algorithms~\cite{ref_soknl, ref_adapi}. 
This section introduces some of the datasets relevant to this work.

Abalone~\cite{ref_abalone} is a well-known dataset from the UCI Machine Learning Repository. It contains measurements of physical attributes of abalones, such as length, diameter, weight, and shell dimensions, aimed at predicting the age of the abalones.
The Bike Sharing dataset~\cite{ref_bike} is a commonly used regression dataset that provides data on bike rental demand, integrating environmental and seasonal settings with temporal features.
The House8L dataset~\cite{ref_house}, derived from the 1990 U.S. Census, is designed for regression tasks aimed at predicting the median house prices in various regions. 
% It comprises 8 input variables, including demographic and socio-economic factors, and is commonly used to evaluate regression models in machine learning research.
The Superconductivity dataset, presented by Hamidieh in~\cite{ref_super}, contains data for predicting the critical temperature of superconductors based on material properties. 
% This dataset serves as a valuable resource for exploring machine learning applications in materials science and advancing the understanding of superconductivity. 
Real-time energy price datasets from~\cite{ref_nzep} capture dynamic electricity pricing and consumption patterns in New Zealand. These datasets include real-time pricing, historical trends, and contextual variables, emphasizing the role of real-time analysis in managing energy demand and supply for sustainability and cost-effectiveness. Consistently updated pricing data ensures relevance for research and application.

Table~\ref{tab:real} summarizes basic information of the real-world datasets, the $\uparrow$ indicates that the instance amount is growing.

\begin{table}[h!]
\centering
\caption{Real-World Dataset Overview}
\label{tab:real}
\begin{tabular}{lccc}
\toprule
\textbf{\textsc{Dataset}}       & \textbf{\textsc{Acronym}}     & \textbf{\textsc{\# Instance}} & \textbf{\textsc{\# Feature}} \\ \midrule
Abalone                & ABA                  & 4,177                & 8                   \\
Bike Sharing           & BIK                  & 17,379               & 12                  \\
House8L                & H8L                  & 22,784               & 8                   \\
Superconductivity      & SUP                  & 21,263               & 82                  \\
NZ Energy Price        & NZEP                 & 34,980$^\uparrow$    & 11                  \\ \bottomrule
\end{tabular}
\end{table}

\section{Problem Definition}
Regressive analysis in stream learning focuses on building and maintaining models capable of predicting continuous target variables from evolving data streams. Unlike traditional batch learning, stream learning must accommodate challenges such as real-time processing, limited memory, and concept drifts. Regression tasks aim to provide accurate point predictions for the target variable, which are crucial for decision-making in dynamic environments. Beyond point predictions, stream learning also emphasizes uncertainty quantification through Prediction Interval (PI), which define a range within which the true target value is expected to lie with a given confidence level. 

\subsection{Evaluation Protocols}
Evaluation under streaming setting is different from conventional machine learning. In addition to the overall performance, the trends and fluctuations during the processing of a stream are also vital \cite{ref_nuwan, ref_moa}. With the drift detection and adaptation abilities, the streaming algorithms ought to recover the performance from the presence of concept drifts, which will not be reflected in the overall evaluations. Thus, in stream learning, researchers usually present two evaluation protocols, \texttt{Cumulative} and \texttt{Prequential} \cite{ref_capymoa}. 

\paragraph{Cumulative Evaluation:}
Cumulative evaluation computes the performance metrics over the entire stream from the beginning up to the current moment. It provides a global measure of the model's performance and is defined as:
\begin{equation}
\label{eq:cumulative_metric}
\text{Metric}_{\text{cumulative}} = \frac{1}{t} \sum_{i=1}^{t} f(y_i, \hat{y}_i),
\end{equation}
where \( t \) is the number of observed data points, \( y_i \) is the actual value, \( \hat{y}_i \) is the predicted value, and \( f(\cdot) \) represents the evaluation function, such as mean absolute error.

\paragraph{Prequential Evaluation:}
Prequential evaluation, also known as interleaved test-then-train, uses a sliding window of the most recent \( n \) data points to compute performance metrics. This approach captures the model's adaptability to recent data and is defined as:
\begin{equation}
\label{eq:prequential_metric}
\text{Metric}_{\text{prequential}} = \frac{1}{n} \sum_{i=t-n+1}^{t} f(y_i, \hat{y}_i).
\end{equation}
Here, \( n \) is the window size, and the summation considers only the most recent \( n \) data points.
% Since prequential evaluation will produce multiple outputs along with the data streams, typically we illustrate them with a plot in a sequential order.

\subsection{Regression}
Literature has proposed plenty of evaluation metrics for regression tasks, most of which are either error-based or correlation-based \cite{ref_r2better}. We recommend one of each in this work.

\subsubsection{Evaluation Metrics}
Two commonly used metrics for evaluating regression models in data streams are:

\paragraph{Root Mean Squared Error (RMSE):}
RMSE measures the average magnitude of errors between predicted and actual values, emphasizing larger errors. It is defined as:
\begin{equation}
\label{eq:rmse}
\sigma_{e} = \text{RMSE} = \sqrt{\frac{1}{n} \sum_{i=1}^{n} (y_i - \hat{y}_i)^2}.
\end{equation}
RMSE is widely used because it provides an interpretable scale of prediction error, matching the units of the target variable.

\paragraph{Adjusted \( R^2 \):}
Adjusted \( R^2 \) evaluates the proportion of variance explained by the model, accounting for the number of predictors to avoid overfitting. It is defined as:
\begin{equation}
\label{eq:adjusted_r2}
\mathcal{R}^2_{adj} = \text{Adjusted } \mathcal{R}^2 = 1 - \left( \frac{(1 - R^2)(n - 1)}{n - p - 1} \right),
\end{equation}
where \( \mathcal{R}^2 \) is calculated as:
\begin{equation}
\label{eq:r2}
\mathcal{R}^2 = 1 - \frac{\sum_{i=1}^{n} (y_i - \hat{y}_i)^2}{\sum_{i=1}^{n} (y_i - \bar{y})^2}.
\end{equation}
Here, \( n \) is the number of data points, \( p \) is the number of predictors, and \( \bar{y} \) is the mean of actual values.

RMSE is selected because it penalizes larger errors, which are critical in regression tasks, especially in applications where large deviations have significant consequences. Adjusted \( R^2 \) is chosen for its ability to measure explanatory power while adjusting for model complexity, ensuring robust evaluation in dynamic data stream scenarios.

\subsection{Prediction Interval}
As presented in~\cite{ref_adapi, ref_conformal}, evaluating prediction intervals is a challenging task. Extremely wide intervals may cover more ground truths but significantly reduce informativeness. Balancing this trade-off requires optimizing two key metrics.

\subsubsection{Evaluation Metrics}

In the literature (\cite{ref_BLM, ref_nmpiw1, ref_nmpiw2,ref_nmpiw3}), two key metrics are commonly used to evaluate the quality of PIs in stream learning:

\paragraph{Coverage}
Coverage measures the percentage of actual target values that fall within the predicted intervals. It evaluates the reliability of the PI and is defined as:
\begin{equation}
\label{eq:coverage}
\mathcal{C} = \text{Coverage} = \frac{1}{n} \sum_{i=1}^{n} \mathbb{I}\left(y_i \in [\hat{y}_i^{L}, \hat{y}_i^{U}]\right),
\end{equation}
where \( n \) is the number of data points, \( y_i \) is the actual value, \( [\hat{y}_i^{L}, \hat{y}_i^{U}] \) represents the lower and upper bounds of the prediction interval for the \( i \)-th observation, and \( \mathbb{I}(\cdot) \) is an indicator function that equals 1 if the condition is satisfied and 0 otherwise.

\paragraph{Normalized Mean Prediction Interval Width (NMPIW)}
NMPIW evaluates the sharpness of the PIs by measuring their average width, normalized by the range of the actual values. It is defined as:
\begin{equation}
\label{eq:nmpiw}
\mathcal{W}_{norm} =\text{NMPIW} = \frac{1}{n R} \sum_{i=1}^{n} \left(\hat{y}_i^{U} - \hat{y}_i^{L}\right),
\end{equation}
where \( R \) is the target values range: \( R = \max(y) - \min(y) \).

The combination of Coverage and NMPIW is essential for a balanced evaluation of PIs. Coverage ensures that the intervals capture the true target values with high reliability, while NMPIW evaluates the precision and sharpness of the intervals. Using both metrics simultaneously ensures that the intervals are neither overly wide (leading to poor precision) nor too narrow (resulting in low reliability). This trade-off is crucial for robust and meaningful uncertainty quantification in dynamic stream learning environments.

\section{Related Works}\label{sec:related}
Regression analyses have always been overlooked in Machine Learning field, not to mention the contempt for the streaming scenario.
Novel research on regression and prediction intervals is limited in the recent literature. 
We choose some new and commonly used algorithms as demonstration tools in this work.

One of the most famous streaming regression models is the Fast Incremental Model Tree with Drift Detection (FIMT-DD)~\cite{ref_fimtdd}. FIMT-DD incrementally constructs regression trees by splitting nodes based on variance reduction, using adaptive sliding windows to detect and respond to concept drift, ensuring timely updates to the model structure.
% The ORTO (Option-enhanced Regression Tree for Online learning)~\cite{ref_orto} algorithm is a Hoeffding-based regression tree method that incorporates decision options to improve predictive performance. By exploring multiple potential splits at decision nodes, ORTO balances accuracy and computational efficiency, enabling faster adaptation to evolving data streams.
The Adaptive Random Forest for Regression (ARF-Reg)~\cite{ref_arfreg} algorithm builds an ensemble of regression trees, leveraging online bagging with weighted resampling and drift detection mechanisms to dynamically adapt individual trees or the entire ensemble to changes in data streams. Noticeably, the based learner of ARF-Reg is typically FIRT-DD, a variant of FIMT-DD. FIRT-DD utilizes mean target values from the leaf as the final prediction instead of a model output to avoid overflow problems.
The Self-Optimising k-Nearest Leaves (SOKNL)~\cite{ref_soknl} algorithm integrates k-nearest neighbors with ARF-Reg, dynamically selecting optimal leaf nodes based on a dissimilarity measurement between centroids in the leaf and the incoming instances. It also adjusts k-values to enhance prediction accuracy on evolving data streams.
Furthermore, a sliding-window k nearest neighbours algorithm is also involved in this work due to its universal usage and surprising effectiveness. 

Recently, there are some developments in the streaming Prediction Interval aspect. Sun et al. implemented a streaming version of Mean and Variance Estimation (MVE) method and proposed a novel Adaptive Prediction Interval (AdaPI)~\cite{ref_adapi} algorithm. The Mean and Variance Estimation (MVE) algorithm is a straightforward method for constructing prediction intervals in regression tasks, operating under the assumption that predictive errors follow a Gaussian distribution. It calculates intervals centered around the predicted value, extending by a factor determined by the inverse Gaussian distribution and the standard deviation of errors. Building upon MVE, the Adaptive Prediction Interval (AdaPI) algorithm introduces a dynamic scaling coefficient that adjusts the interval width based on historical coverage. This adaptive mechanism ensures that the prediction intervals converge towards a user-defined confidence level over time, making AdaPI particularly suitable for streaming data where the underlying data distribution may evolve.

\section{Data Simulation}

\subsection{Augmenting Real World Datasets}
As exhibited in Section~\ref{sec:real-data}, the real-world datasets for regression usually consist of ``insufficient'' instances. However, in order to perform research according to the streaming protocol, the datasets ought to be at substantial length. Therefore, we leverage the development of the generative model and utilize Generative Adversarial Networks (GANs)~\cite{ref_gans} to enhance the real-world datasets. 

In particular, the real-world datasets were used as the input data source to the Conditional Tabular GANs (CTGANs)~\cite{ref_ctgan} from the Synthetic Data Vault (SDV)~\cite{ref_sdv}. With the trained generative model, unlimited amount of data points are available for further use. 

In this work, the following parametrization is specified to all generative models:
\begin{inparaenum}
    \item epoch number: 300;
    \item batch size: 500; and
    \item learning rate for generator and the discriminator: 0.001.
    % \item learning rate for discriminator: 0.001.
\end{inparaenum}
The other parameters conform with the default values in SDV. All the parameters can be conveniently tuned in our scripts.

\subsection{Synthesizing Concept Drifts}
In practice, recognizing and defining concept drifts are quite tricky~\cite{ref_moa, ref_insect}.
Without a distinct definition of concept drifts, simulating them is as well an unclear task. 

Particularly, the simulation of incremental drift has been challenging the streaming researchers for over a decade.
In the current literature, there are two commonly used incremental simulator: Hyperplane~\cite{ref_hyper} and Radial Basis Function (RBF)~\cite{ref_rbf1, ref_rbf2}. Both of them define the concept with their geometric properties. In the Hyperplane case, it is the flat surface in the high-dimensional space, and RBFs rely on the random centroids in the input space.
The incremental drifts are simulated by the constant movements of the plane or centroids~\cite{ref_srp, ref_sgbt}.

In~\cite{ref_insect}, researchers selected the temperature at the moment of the data collection as the drifting feature. Inspired by this idea, we propose a concept simulation approach that utilizes one of the numeric features from the datasets as the concept defining feature, and exploit drift synthesizing methods on the feature to form different types of drifts.

Initially, a ``drifting feature'' is selected. We apply a pair-wise correlation test between the target and each feature column on the generated data. Multiple methods are available from statistical and machine learning fields, such as pearson test~\cite{ref_pearson}, spearman test~\cite{ref_spearman}, feature importance~\cite{ref_featureimportance}, and so forth. Only numeric features can be candidates as the drifting feature for two reasons: (1) numeric values provides more informativeness than categorical ones; and (2) only with numeric values can we synthesize incremental drifts in the following research.

Next, the datasets are sorted based on the drifting feature and separated into distinct chunks. The number of chunks are decided according to the amount of required concepts. All chunks will have different central values in terms of the drifting feature. Each particular chunk will be used for training a CTGAN, which will be later leveraged as the source of a specific concept. 

\subsubsection{Abrupt Drift}
Simulating abrupt drifts is relatively straightforward. 
With a specific number of drifts, a proper amount of concepts can also be determined. A data stream with a desired length is generated using the associated CTGAN for each concept. These streams will later be vertically concatenated together in a random order (to avoid drifting trends).

\subsubsection{Gradual Drift}
The simulation procedure of a gradual drift is a bit more complex.
we employ the following procedure:
\begin{enumerate}
    \item Extract the last \( n \) instances from the initial concept (\( C_1 \)) and the first \( n \) instances from the subsequent concept (\( C_2 \)).
    \item Combine these \( 2n \) instances from \( C_1 \) and \( C_2 \) into a single set and apply random shuffling to intermix the instances to construct a drifting period.
    \item Construct the final data stream by concatenating \( C_1 \), the shuffled \( 2n \)-instance drift segment, and \( C_2 \).
\end{enumerate}
This process generates a smooth transition between \( C_1 \) and \( C_2 \), simulating a gradual drift in the underlying data distribution and repeats until the data stream reaches the desired amount of drifts.

\subsubsection{Incremental Drift}
The steps for synthesizing incremental drift are similar to gradual ones. However, in step (2), instead of shuffling, the data is sorted based on the values in the drifting feature. The direction of the ordering is determined by the average value difference between the concepts at both ends of the drifting period.
To avoid potential information leakage caused by the sorted values in the drifting features, in our work, the drifting feature will be discarded from the datasets.
Figure~\ref{fig:teaser} plots the drifting feature and the target values from the original Bike datasets after simulating two incremental drifts with 4000 drift length. 
Evidently, the drifting feature is arranged in an incremental manner. A similar trend can be faintly observed in the target values, although it is not very prominent. This ensures that the simulated concept drift indeed exists while being difficult to detect straightforwardly.
% Specifically, the feature and target values are just random values of [0,10] and [-2,2], respectively. 
% Evidently, even the drifting feature shows obvious patterns, similar trends can not necessarily be observed in the target values.

\subsubsection{Special Case for NZEP}
We treated NZEP dataset group differently in this work due to the extra information they provides.
Because the price data is collected from scattered locations across New Zealand, the location information can spontaneously distinguish different concepts. Furthermore, the horizon (how many steps ahead of the forecasting) is also a vital factor when abstracting knowledge from the data. Please see the original work~\cite{ref_nzep} for more details.
Thus, when simulating abrupt and gradual drifts on the NZEP data, we choose locations along with different horizons as the concept. Unfortunately, incremental drifts cannot be synthesized on them since these two features, to some degree, also represent categorical information.

\subsection{Synthesized Datasets}
To demonstrate the outcomes of our work, we present 18 synthesized datasets and list the details in this section.

Please note that since there are countless combinations of the locations and horizons for NZEP datasets, we manually selected four representatives: 
\begin{inparaenum}
    \item Auckland, with 4 hours horizon; 
    \item Hamilton, with 6 hours horizon; 
    \item Dunedin, with 30 minutes horizon; and 
    \item Wellington, with 24 hours horizon.
\end{inparaenum}

The generation of the datasets is detailed as follows:
In general, for abrupt drifts, we created four concepts with a length of 50k with the source data. As a consequence, all the ``abrupt'' datasets contains 200k instances. 
In terms of gradual drifts, the concepts are also 50k instances long and the drifting periods are uniformly set to 10k. 
As afore-explained, incremental drifts are simulated with the entire data source instead of parts of them, the CTGANs associated with incremental drifts are separately trained. The generated incremental datasets have 100k instances, and 2 incremental drifts with drifting period of 20k. In this manner, the datasets are evenly sectored into 5 parts -- 3 stable periods and 2 drifting periods.
Each location for NZEP data is individually simulated a incremental dataset, and named after the city, i.e., ALK, HAM, WEL, and DUN.
Table~\ref{tab:syn-data} illustrates overview of the generated datasets. 
% Note that drift types are represented with their initials.
% Drift Types -- A: Abrupt; G: Gradual; I: Incremental
\begin{table}[!htbp]
\captionsetup{justification=centering}
\centering
\caption{Overview of Synthetic Datasets \\ Drift Type Notation -- A: Abrupt; G: Gradual; I: Incremental}
\label{tab:syn-data}
\begin{tabular}{ccccc}
\toprule
\textbf{\textsc{Datasets}} & \textbf{\textsc{\# Instance}} & \textbf{\textsc{\# Feature}} & \textbf{\textsc{\# Drifts}} & \textbf{\textsc{Type}} \\
\midrule
ABA$_{3a}$ & 200,000 & 8  & 3  & A \\
ABA$_{3g}$ & 200,000 & 8  & 3  & G \\
ABA$_{2i}$ & 100,000 & 7  & 2  & I \\\hline
BIK$_{3a}$ & 200,000 & 12 & 3 & A \\
BIK$_{3g}$ & 200,000 & 12 & 3 & G \\
BIK$_{2i}$ & 100,000 & 11 & 2 & I \\\hline
H8L$_{3a}$ & 200,000 & 8  & 3  & A \\
H8L$_{3g}$ & 200,000 & 8  & 3  & G \\
H8L$_{2i}$ & 100,000 & 7  & 2  & I \\\hline
SUP$_{3a}$ & 200,000 & 82 & 3 & A \\
SUP$_{3g}$ & 200,000 & 82 & 3 & G \\
SUP$_{2i}$ & 100,000 & 81 & 2 & I \\\hline
NZEP$_{3a}$ & 200,000 & 11 & 3 & A \\
NZEP$_{3g}$ & 200,000 & 11 & 3 & G \\
AKL         & 100,000 & 10 & 2 & I \\
HAM         & 100,000 & 10 & 2 & I \\
WEL         & 100,000 & 10 & 2 & I \\
DUN         & 100,000 & 10 & 2 & I \\
\bottomrule
\end{tabular}

\end{table}

% ########################### old ideas
% Where we provide script code for simulating drifts with single or multiple input datasets.
% Similar to the insect datasets, we can take one dataset, select a feature that makes semantical sense to split the dataset, and we reorder the dataset based on the feature to simulate drifts.

% We can provide some functions to simulate sudden, gradual, or incremental. 
% We can also simulate recurring drifts with different drifting manners.

% {\color{red} 
% Sudden: append a concept after another one directly;

% Gradual: take the last $n$ instances from the first concept, and the take the first $n$ instances from the second concept. merge this $2n$ instances and shuffle them. And then concatenate the first concept, the merged drifting $2n$ instances, and the second concept.

% Incremental: similar approach to gradual to acquire $2n$ instances. order them with a use-specified feature (or other manners). Concatenation.

% Recurring: as it should be
% }
\begin{figure*}[!htbp]
    \centering
    \includegraphics[width=.95\linewidth]{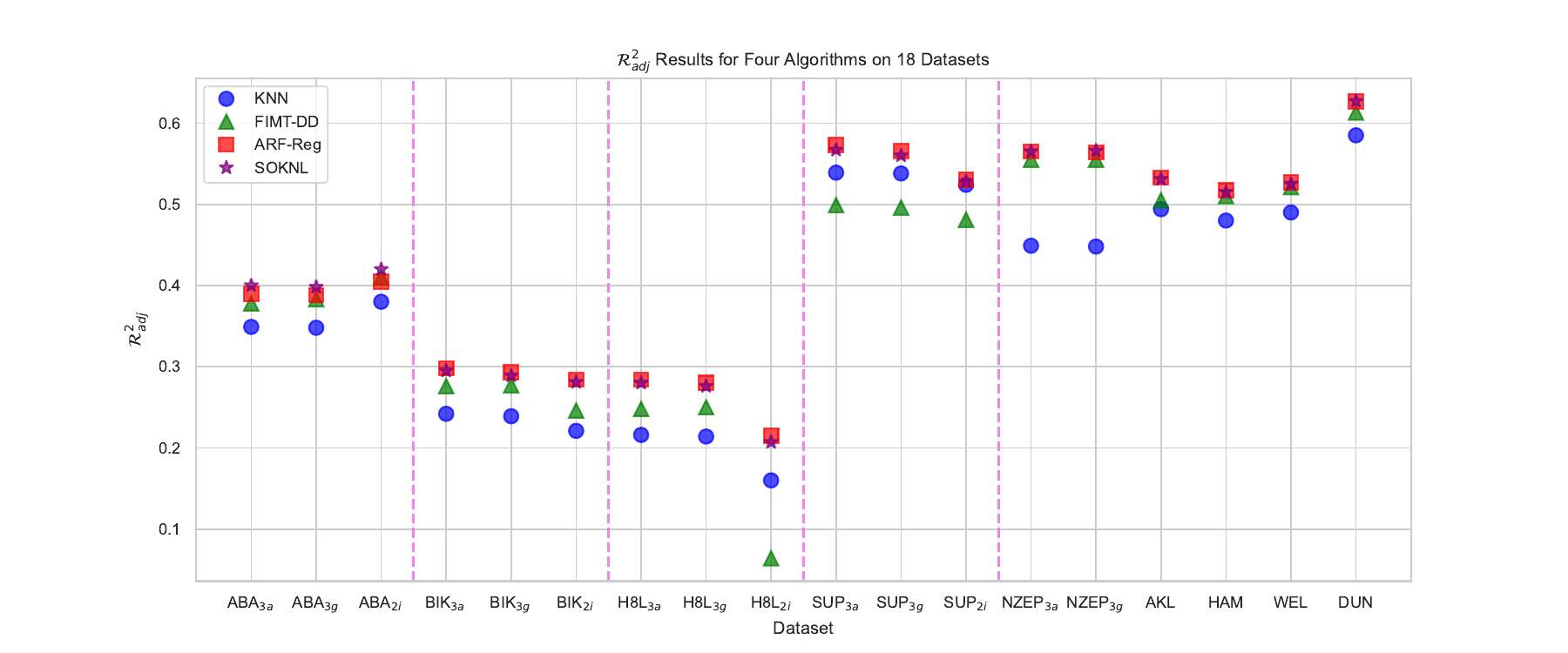}
    \caption{Adjusted R-squared ($\mathcal{R}^2_{adj}$) Results for Four Algorithms on 18 Datasets}
    \label{fig:r2_fig}
\end{figure*}

\section{Experiments and Discussion}
This section introduces the conducted experiments, exhibits the results, and facilitates associated discussions.

\subsection{Algorithms and Parametrization}
The following configurations are provided to ensure and reproducibility:
\begin{inparaenum}
    \item \textbf{Sliding Window KNN}: \( k = 10 \), window size = 1000.
    \item \textbf{FIMT-DD}: Grace period = 200, split confidence = 0.01.
    \item \textbf{ARF-Reg} and \textbf{SOKNL}: Ensemble size = 30.
\end{inparaenum}
These settings align with common literature practices and ensure effective performance across datasets.
For Prediction Interval experiments, both MVE and AdaPI (Section~\ref{sec:related}) used a 95\% confidence level. AdaPI’s lower limit was set to 0.01, while KNN and SOKNL served as base models with the same configurations. 
Prequential executions used a window size of 1000. All participants can be easily accessed from open-sourced stream learning platforms, such as MOA~\cite{ref_moa} and CapyMOA~\cite{ref_capymoa}.
% , two open-sourced stream learning platforms.
% , both of them are open-sourced stream learning platforms.

\subsection{Regression}

Figure~\ref{fig:r2_fig} compares four regression algorithms -- KNN, FIMT-DD, ARF-Reg, and SOKNL -- across 18 datasets using their cumulative Adjusted \( \mathcal{R}^2 \) (\( \mathcal{R}^2_{adj} \)). 
The x-axis represents the datasets, separated into groups by the vertical pink dashed lines. Each group has the same source of original data, e.g., the first group is derived from Abalone dataset.
The standard deviations from ten executions are significantly small, thus omitted in the figure. 
% Since RMSE (\(\sigma_e\)) results from different datasets have different scales,
Because different datasets yield different scale of RMSE (\(\sigma_e\)) results, 
it is unfeasible to illustrate all \(\sigma_e\) results within a single figure without rescaling. Therefore, we avoid the illustration for RMSE in the main contents of this paper. 
Full \( \mathcal{R}^2_{adj} \) and \(\sigma_e\) results can be located in Table~\ref{tab:regression} in Appendix~\ref{apx:reg}.

% SOKNL consistently achieves the best performance, with the highest \( \mathcal{R}^2_{adj} \), demonstrating its accuracy and robustness. ARF-Reg closely follows, making it a competitive alternative for dynamic environments. FIMT-DD and KNN perform less effectively, with KNN showing the weakest results overall. These findings highlight SOKNL and ARF-Reg as reliable choices for regression tasks on streaming data, with SOKNL excelling in both stability and predictive power.

SOKNL and ARF-Reg exhibit comparable performance, both achieving high \( \mathcal{R}^2_{adj} \) values, indicating their accuracy and robustness in streaming regression tasks. While FIMT-DD and KNN perform less effectively, with KNN showing the weakest results overall, the findings suggest that both SOKNL and ARF-Reg are reliable choices for dynamic environments. SOKNL maintains strong stability, while ARF-Reg remains a competitive alternative.

\begin{figure*}[!htp]
    \centering
    % First image with subcaption
    \begin{subfigure}{0.32\textwidth}
        \centering
        \includegraphics[width=\textwidth]{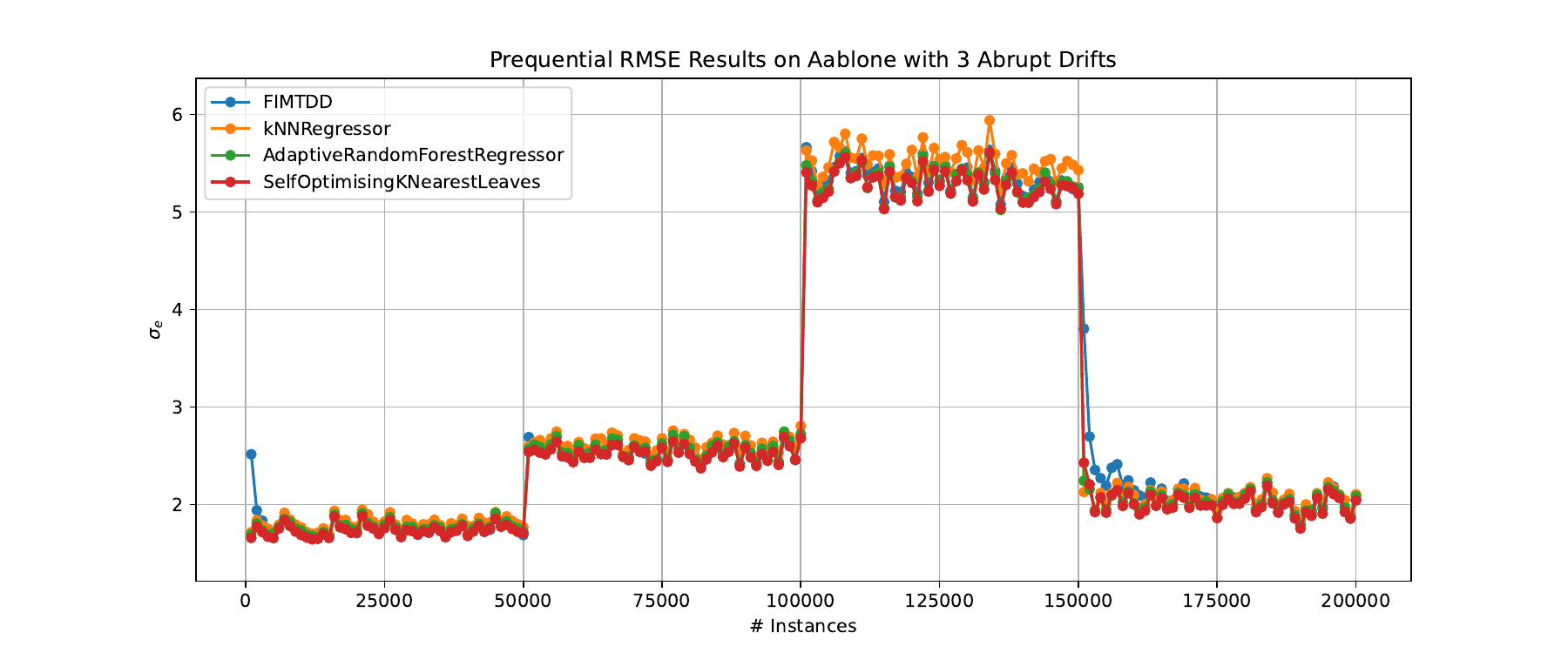}
        \caption{RMSE on Abalone Abrupt}
        \label{fig:aba_a}
    \end{subfigure}
    \hfill
    % Second image with subcaption
    \begin{subfigure}{0.32\textwidth}
        \centering
        \includegraphics[width=\textwidth]{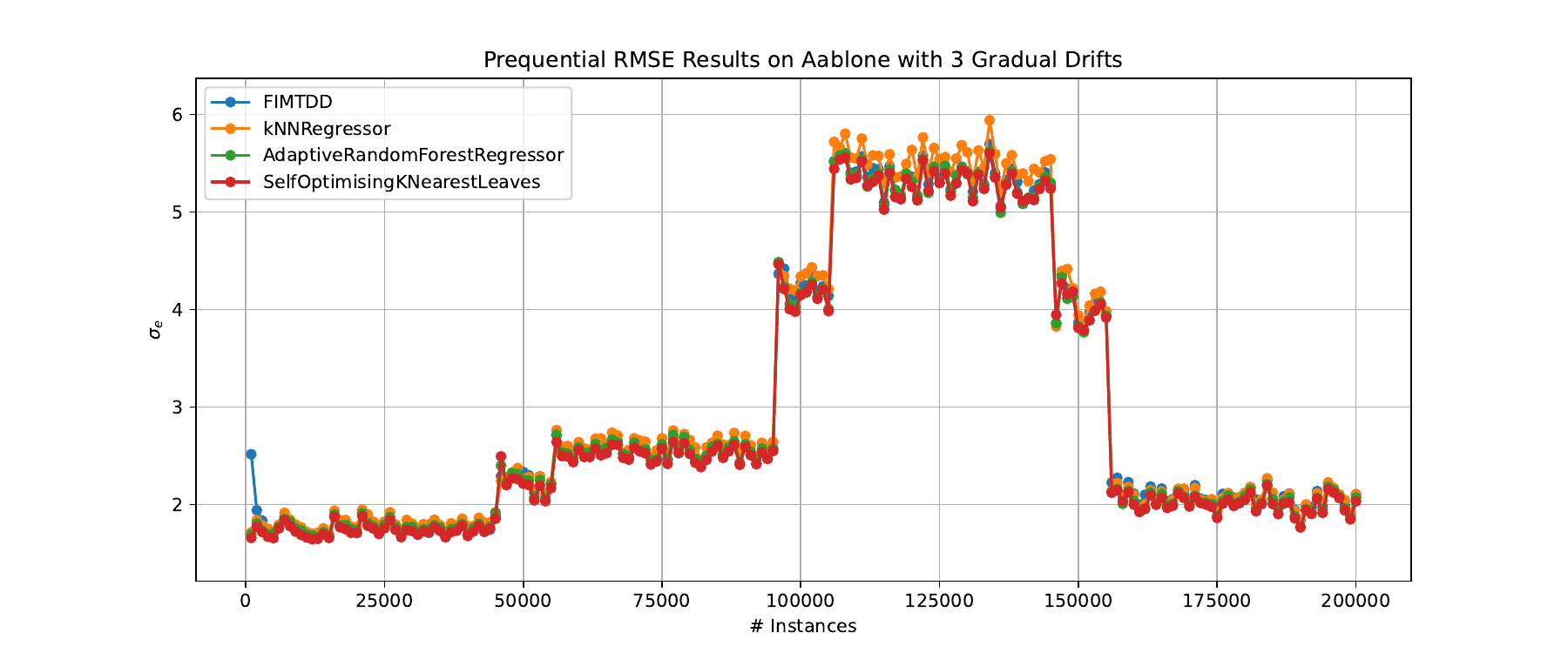}
        \caption{RMSE on Abalone Gradual}
        \label{fig:aba_g}
    \end{subfigure}
    \hfill
    % Third image with subcaption
    \begin{subfigure}{0.32\textwidth}
        \centering
        \includegraphics[width=\textwidth]{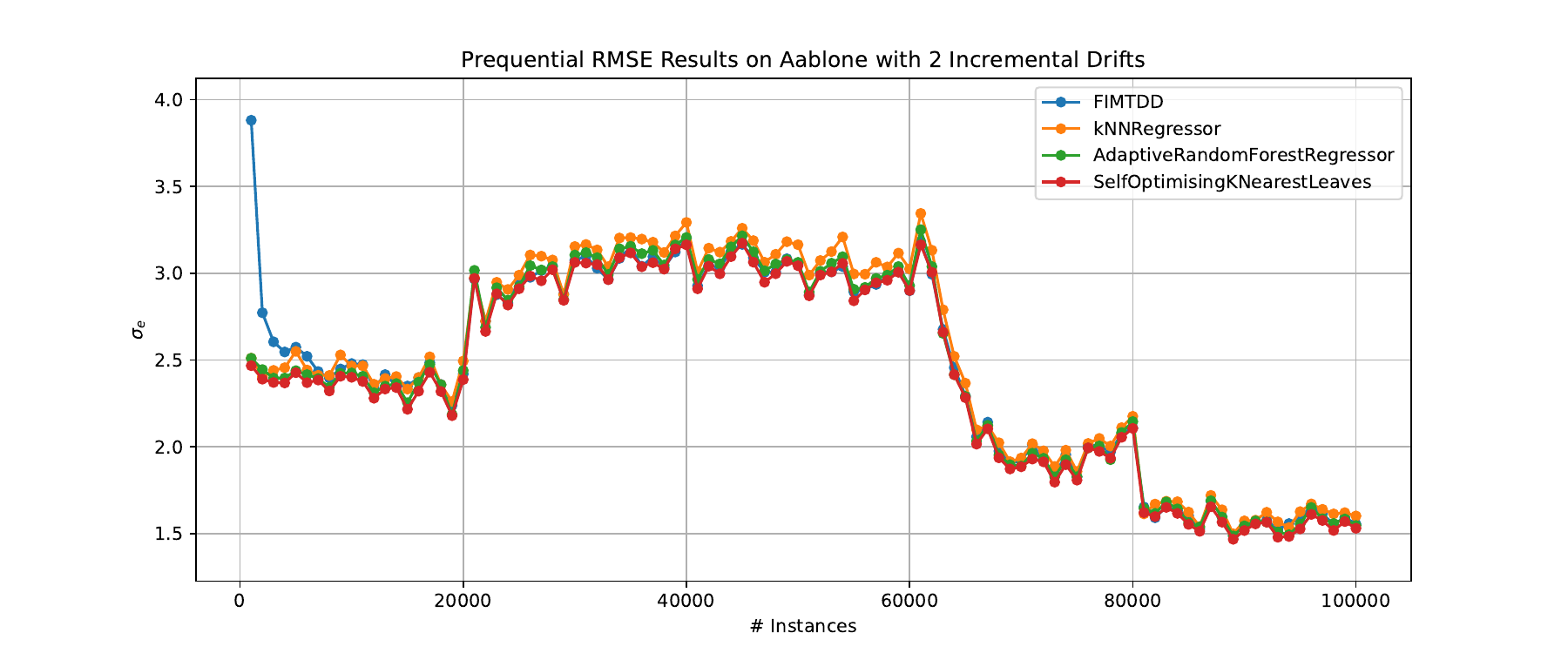}
        \caption{RMSE on Abalone Incremental}
        \label{fig:aba_i}
    \end{subfigure}
    \caption{Prequential RMSE (\(\sigma_e\)) Results for Abalone Dataset Group}
    \label{fig:abalone}
\end{figure*}

Prequential plots show a more streaming perspective of the results.
Due to space restraint, we only illustrate a part of the outcomes here.
Figure~\ref{fig:abalone} exhibits the RMSE results on the Abalone dataset group. 
Apparent sectors can be observed on the ``abrupt-drifted'' dataset (Figure~\ref{fig:aba_a}), while the Figure~\ref{fig:aba_g} shows more fluctuations during the drifting periods.
The incremental one (Figure~\ref{fig:aba_i}) is more turbulent, affirming the effectiveness of the proposed incremental simulation method.

\begin{figure*}[h!]
    \centering
    % First row of images
    \begin{subfigure}{0.32\textwidth}
        \includegraphics[width=\textwidth]{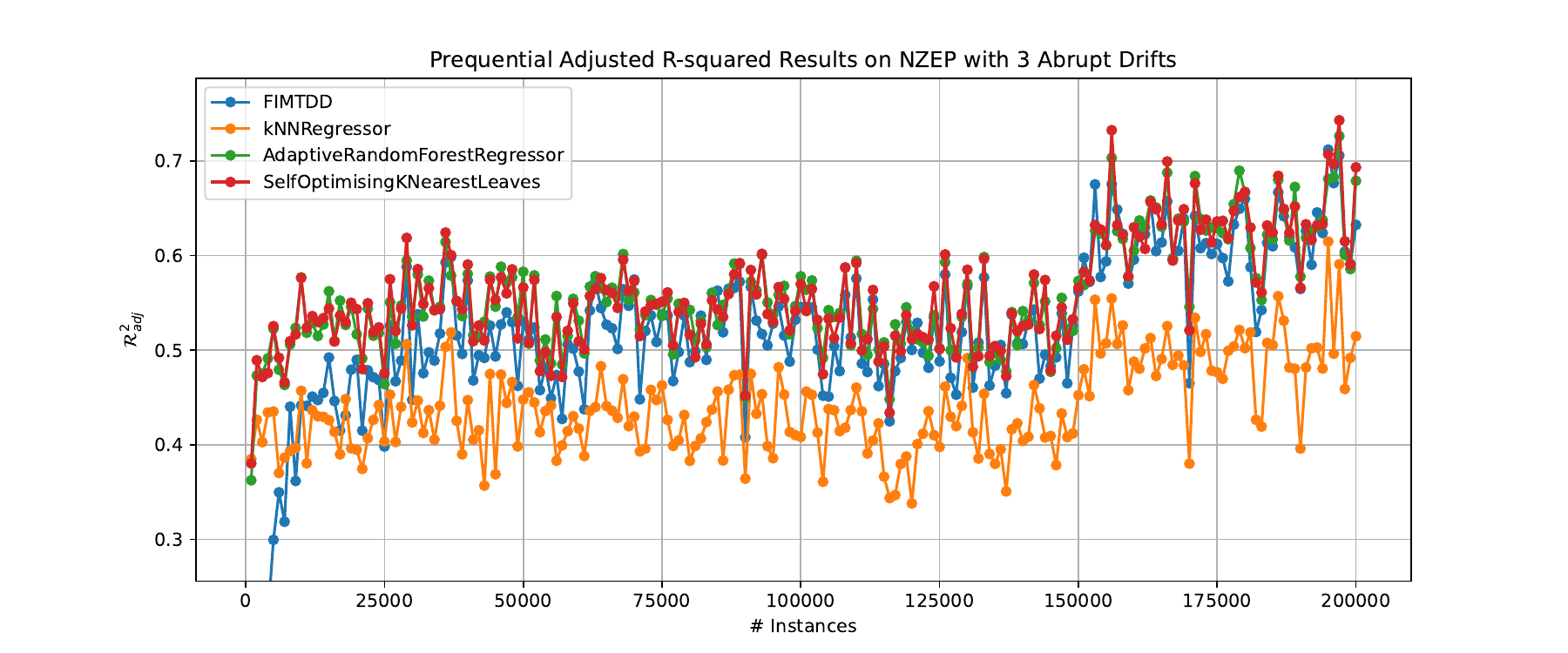}
        \caption{\(\mathcal{R}^2_{adj}\) for NZEP Abrupt}
        \label{fig:nzep_a}
    \end{subfigure}
    \hfill
    \begin{subfigure}{0.32\textwidth}
        \includegraphics[width=\textwidth]{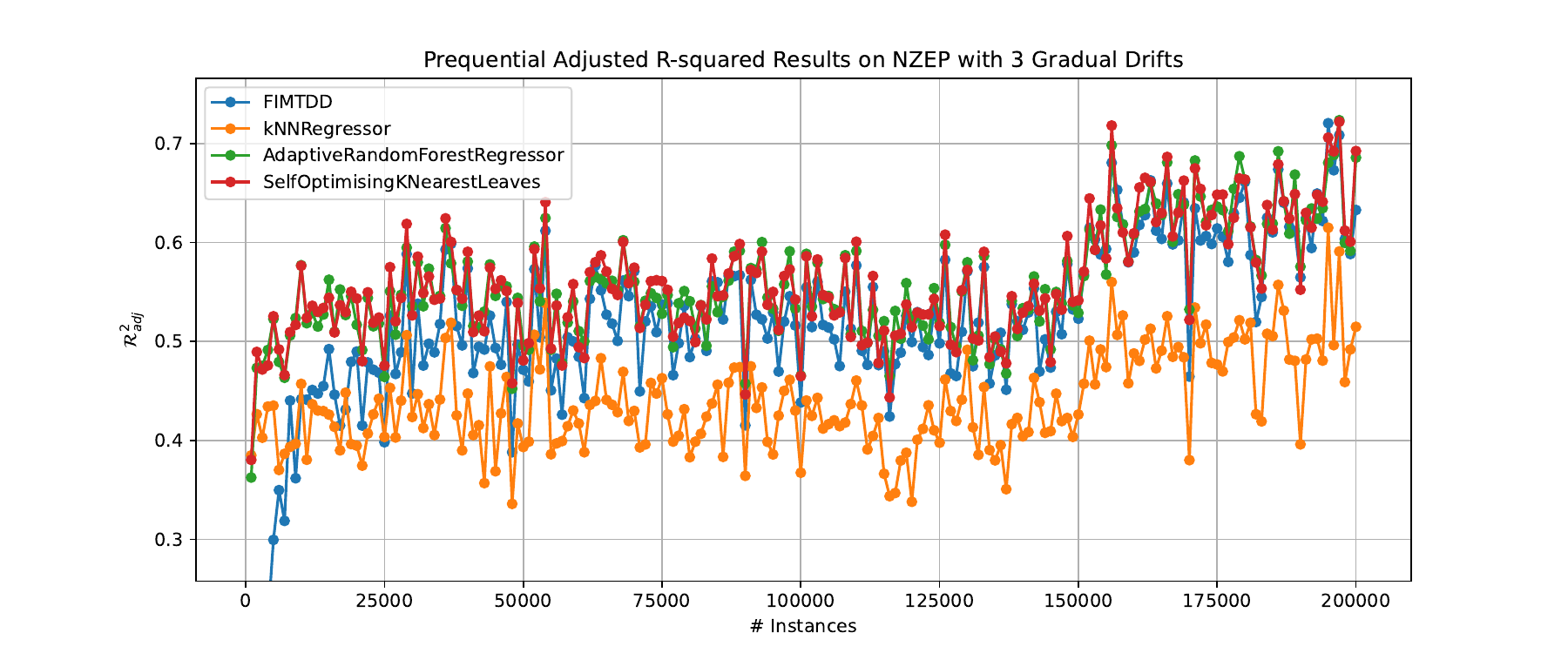}
        \caption{\(\mathcal{R}^2_{adj}\) for NZEP Gradual}
        \label{fig:nzep_g}
    \end{subfigure}
    \hfill
    \begin{subfigure}{0.32\textwidth}
        \includegraphics[width=\textwidth]{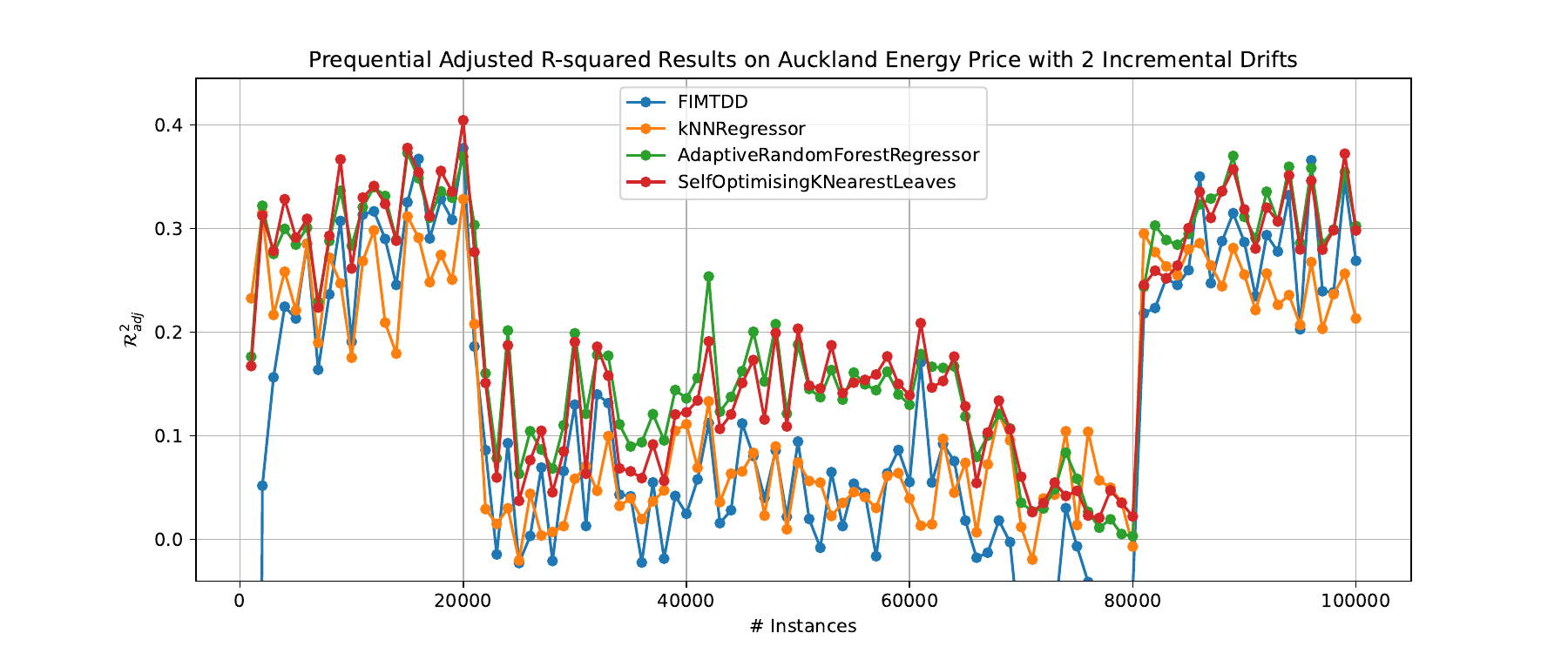}
        \caption{\(\mathcal{R}^2_{adj}\) for Auckland Incremental}
        \label{fig:akl}
    \end{subfigure} \\[1em]
    % Second row of images
    \begin{subfigure}{0.32\textwidth}
        \includegraphics[width=\textwidth]{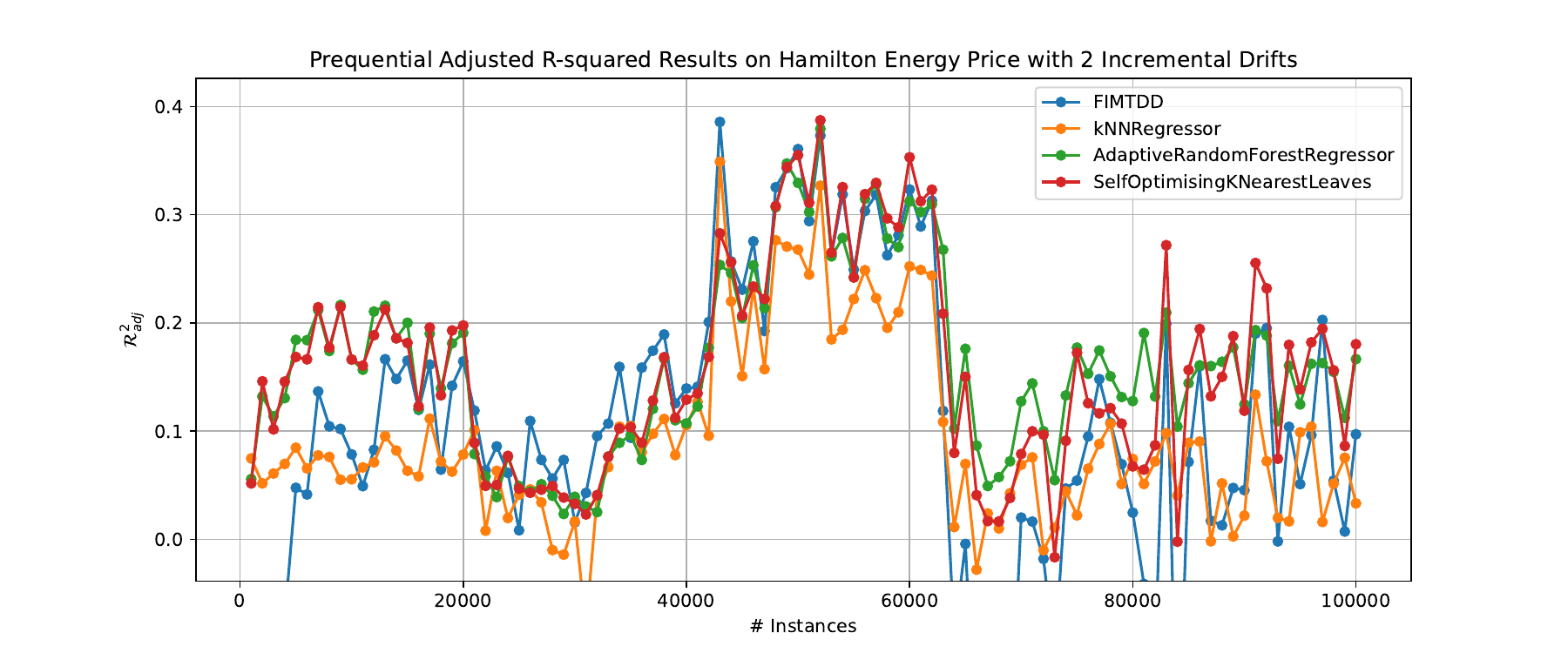}
        \caption{\(\mathcal{R}^2_{adj}\) for Hamilton Incremental}
        \label{fig:ham}
    \end{subfigure}
    \hfill
    \begin{subfigure}{0.32\textwidth}
        \includegraphics[width=\textwidth]{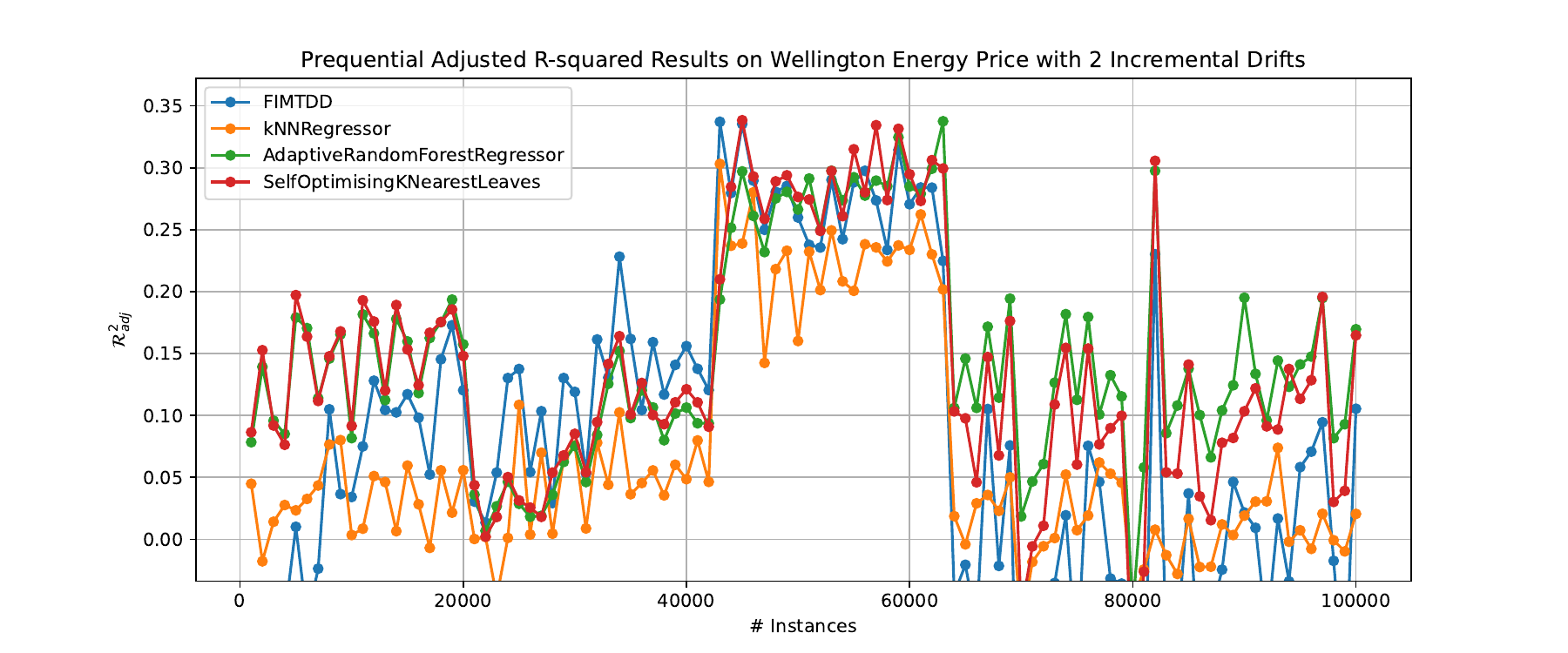}
        \caption{\(\mathcal{R}^2_{adj}\) for Wellington Incremental}
        \label{fig:wel}
    \end{subfigure}
    \hfill
    \begin{subfigure}{0.32\textwidth}
        \includegraphics[width=\textwidth]{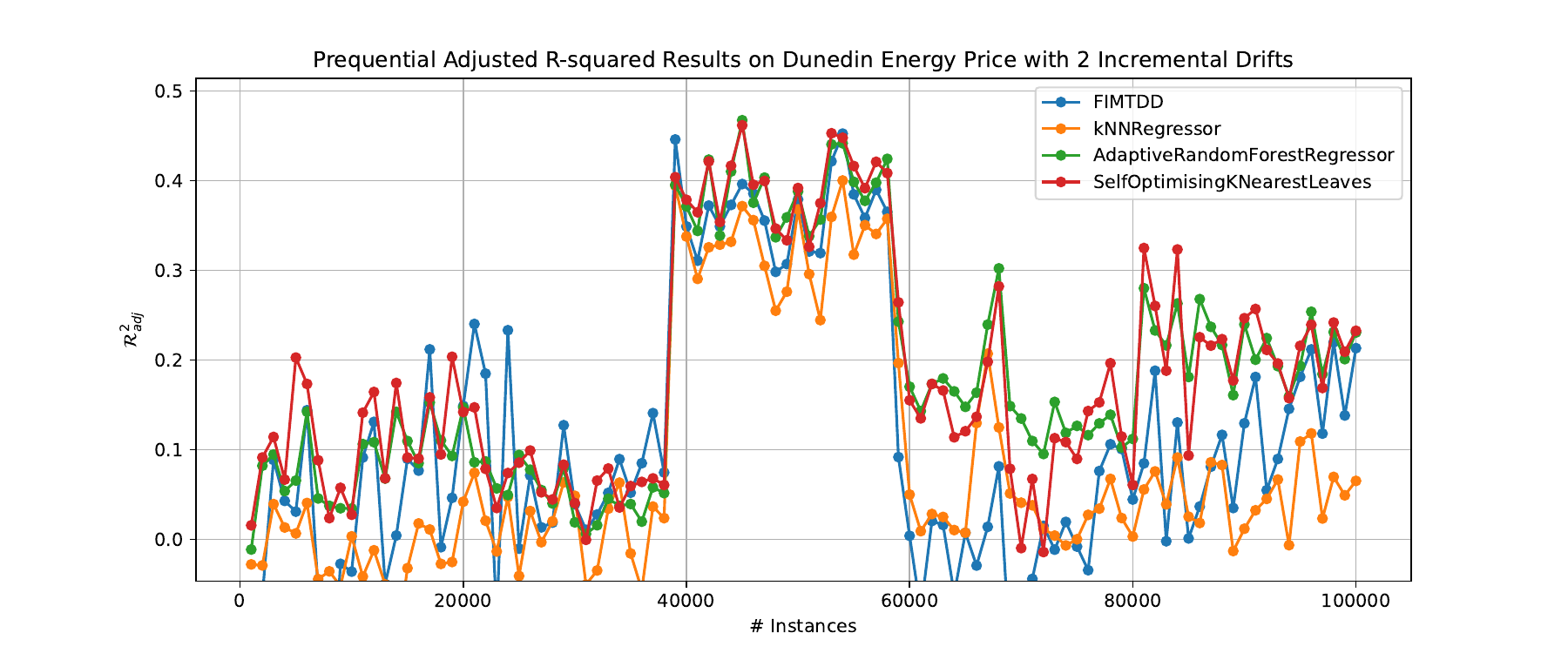}
        \caption{\(\mathcal{R}^2_{adj}\) for Dunedin Incremental}
        \label{fig:dun}
    \end{subfigure}
    \caption{Prequential Adjusted R-squared (\(\mathcal{R}^2_{adj}\)) Results for NZEP Dataset Group}
    \label{fig:nzep}
\end{figure*}

Figure~\ref{fig:nzep} is a prequential \(\mathcal{R}^2_{adj}\) result demonstration for the NZEP dataset group.
A key observation is that for the abrupt and gradual (Figure~\ref{fig:nzep_a} and~\ref{fig:nzep_g}), only the last concept cause an significant difference. In our case, the fourth concept represents the energy price in Dunedin, which is the only city from the South Island of New Zealand. This indicates a potential difference in the pricing strategy between North (AKL, HAM, and WEL) and South Island (DUN).
In terms of the incremental ones (Figure~\ref{fig:akl} --~\ref{fig:dun}), 
in all of which different degrees of unstability can be observed. The reason is likely to be the effectiveness of the incremental drifts simulated with the most correlated feature to the target values.
% they all illustrate different degrees of unstability due to the incremental simulated with the most correlated feature to the target values.

\subsection{Prediction Interval}

\begin{figure*}[!htbp]
    \centering
    \includegraphics[width=.95\linewidth]{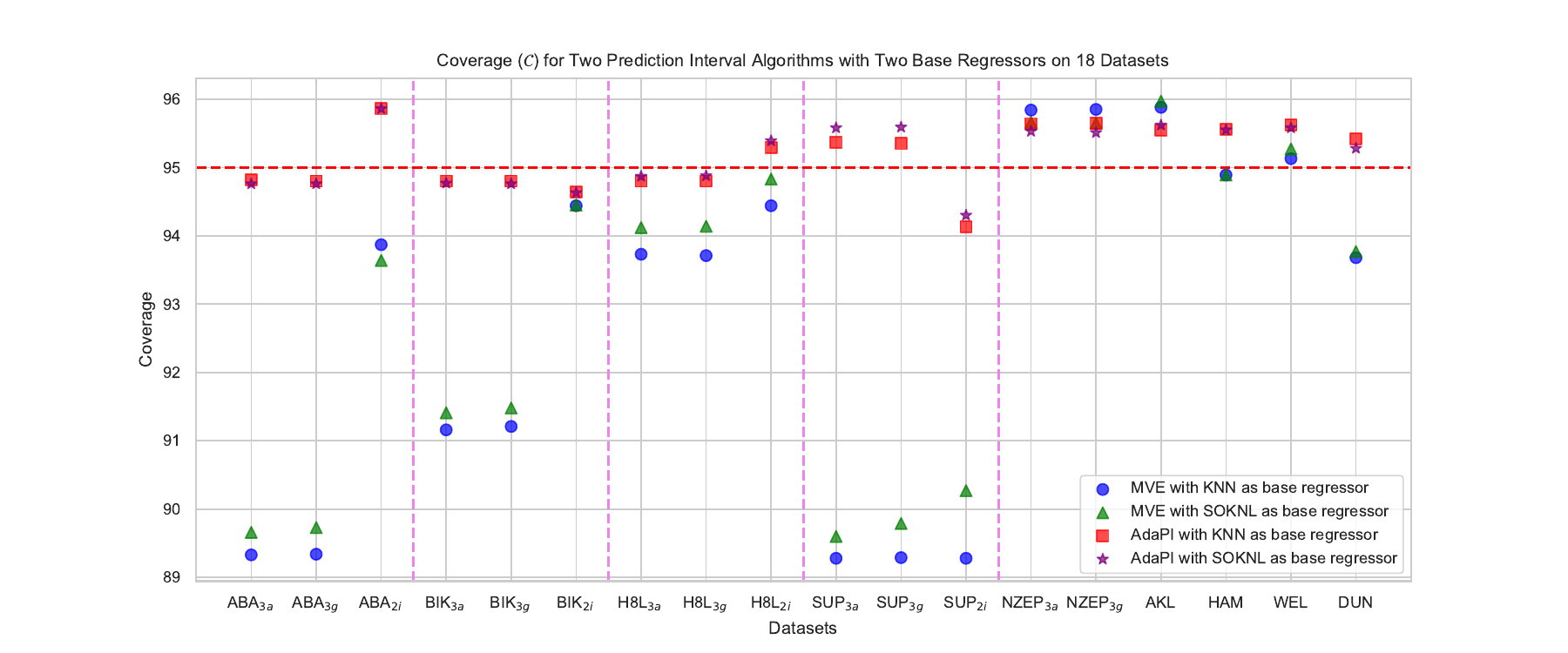}
    \caption{ Coverage ($\mathcal{C}$) for Two Prediction Interval Algorithms with Two Base Regressors on 18 Datasets. The red dashed line highlights the confidence level, which the PI methods aim to be closer to}
    \label{fig:coverage_fig}
\end{figure*}

\begin{figure*}[!htbp]
    \centering
    \includegraphics[width=.95\linewidth]{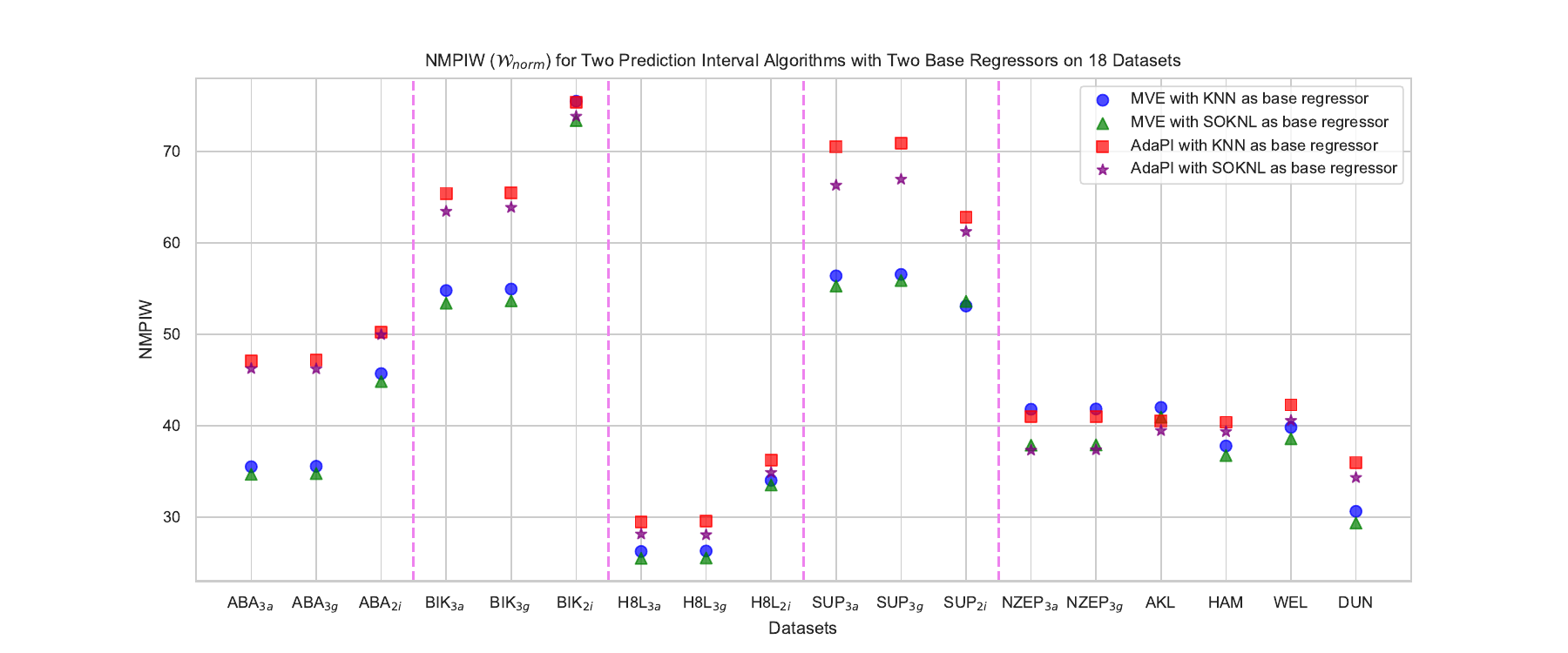}
    \caption{NMPIW ($\mathcal{W}_{norm}$) for Two Prediction Interval Algorithms with Two Base Regressors on 18 Datasets}
    \label{fig:nmpiw_fig}
\end{figure*}

Figure~\ref{fig:coverage_fig} and~\ref{fig:nmpiw_fig} present the cumulative coverage (\( \mathcal{C} \)) and NMPIW (\( \mathcal{W}_{norm} \)) results for different datasets, comparing the performance of MVE and AdaPI methods with KNN and SOKNL models.
A tabular result summarization of the cumulative \( \mathcal{C} \) and \( \mathcal{W}_{norm} \) is situated at Table~\ref{tab:cumulativepi} in the Appendix~\ref{apx:pi}.

Across most datasets, AdaPI consistently achieves higher coverage (\( \mathcal{C} \)) than MVE, indicating more reliable prediction intervals at a 95\% confidence level. However, this increased reliability comes at the cost of wider intervals (\( \mathcal{W}_{norm} \)), as seen in datasets like ABA$_{3a}$ and SUP$_{3a}$. The SOKNL model generally performs better than KNN in terms of narrower prediction intervals (\( \mathcal{W}_{norm} \)) while maintaining competitive coverage (\( \mathcal{C} \)). Notably, the NZEP datasets exhibit excellent coverage and relatively low interval widths, suggesting these methods perform well on energy pricing data. Overall, the results highlight the trade-off between interval width and coverage, with AdaPI and SOKNL offering a balanced and reliable approach.

Figure~\ref{fig:c_w} illustrates the prequential Coverage and NMPIW sequences for MVE and AdaPI (with KNN as the base regressor) on the House8L Abrupt (H8L$_{3a}$) dataset. Three drifts are clearly identifiable, with the second drift causing the most significant impact. Using this as an example, we can analyze the following behavior: 

\begin{figure}[!ht]
    \centering
    \begin{subfigure}{0.45\textwidth}
        \centering
        \includegraphics[width=\textwidth]{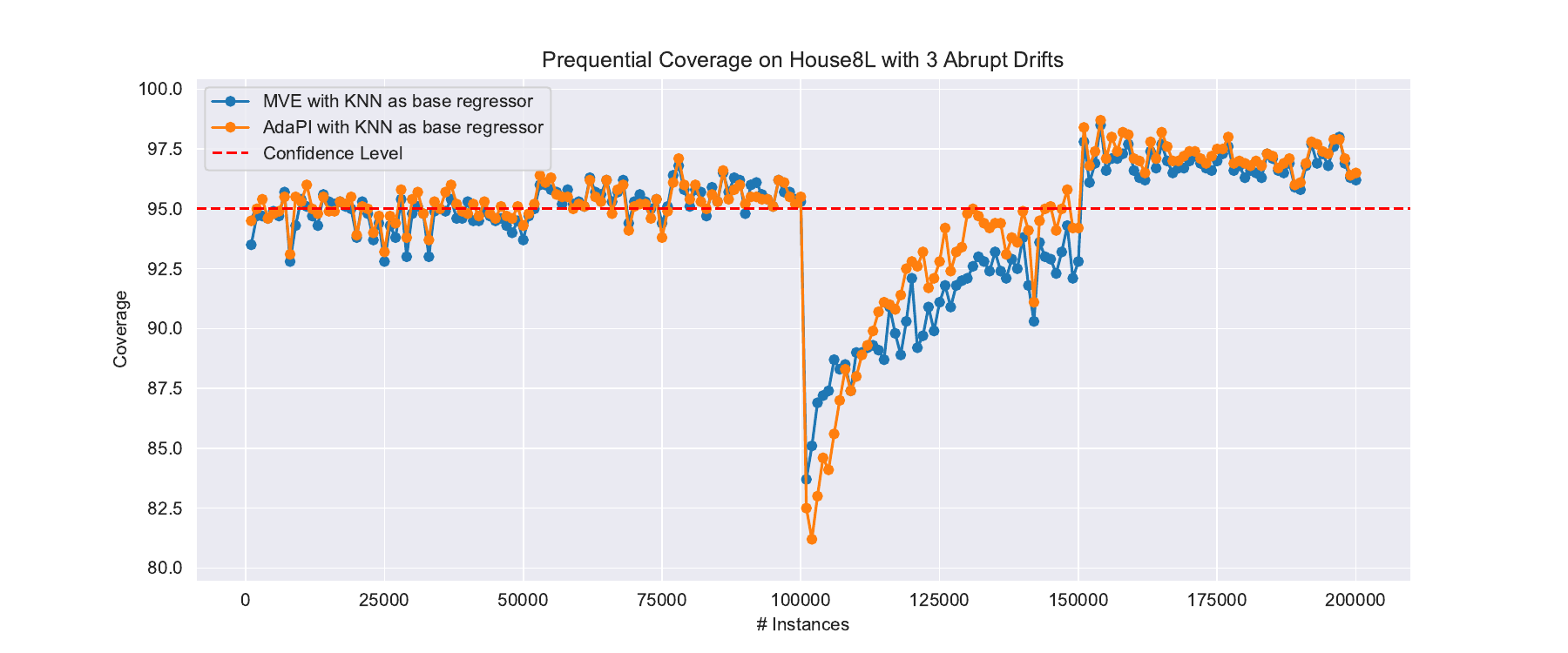}
        \caption{Prequential Coverage (Confidence Level as Red Dashed Line)}
        \label{fig:coverage}
    \end{subfigure}
    % \vspace{0.5cm} % Adjust the spacing between subfigures if necessary
    \begin{subfigure}{0.45\textwidth}
        \centering
        \includegraphics[width=\textwidth]{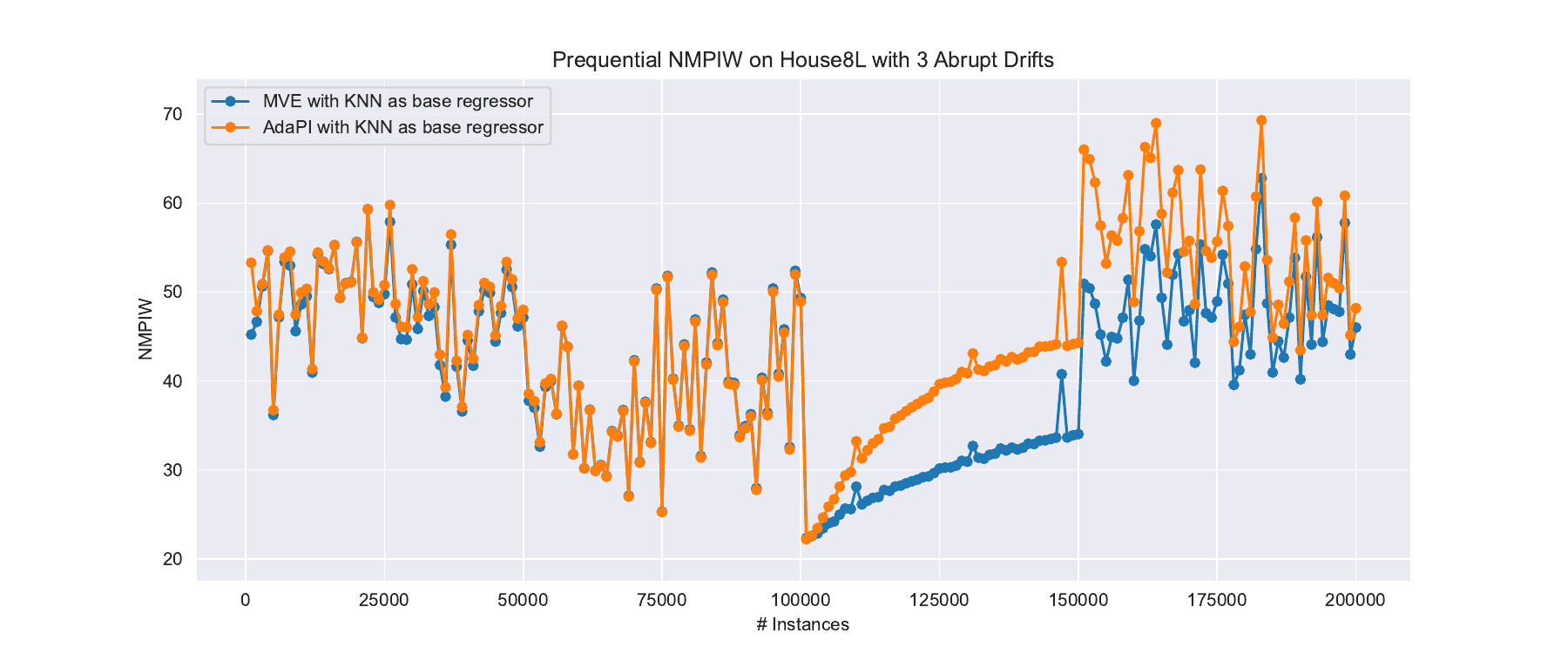}
        \caption{Prequential NMPIW}
        \label{fig:nmpiw}
    \end{subfigure}
    \caption{Prequential Coverage (\(\mathcal{C}\)) and NMPIW (\(\mathcal{W}_{norm}\)) for MVE and AdaPI with KNN as the base regressor on H8L\(_{3a}\)}
    \label{fig:c_w}
\end{figure}

After encountering the second drift, the KNN struggles to represent the new concept, leading to an increase in errors. In response, AdaPI attempts to recover Coverage by widening the generated intervals to capture more ground truths. Consequently, the NMPIW for AdaPI becomes noticeably larger than that of MVE, as depicted in Figure~\ref{fig:nmpiw}. The outcome is reflected in Figure~\ref{fig:coverage}, where the Coverage for AdaPI regains the 95\% at a faster pace.

\begin{figure}[!ht]
    \centering
    \includegraphics[width=\linewidth]{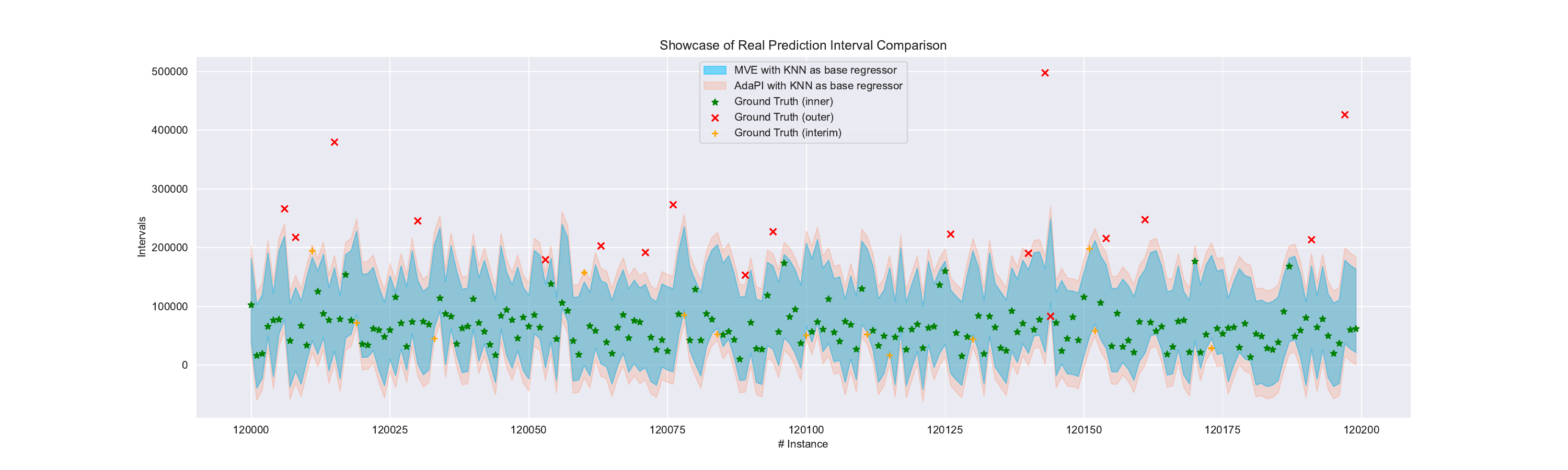}
    \caption{A Clip of the Prediction Interval from MVE and AdaPI after the Second Drift (Instance \#120,000 to \#120,200)}
    \label{fig:real-pi}
\end{figure}

If we take a more focus look into the streaming progress, as demonstrated in Figure~\ref{fig:real-pi}, which is a clip of the entire stream of the generated PI areas, it can be spotted that AdaPI (red shadowed region) is larger than the MVE (blue shadowed region) and covers more target values (the additional truths covered by AdaPI are marked by orange ``+'' symbols). This behaviour makes sure that the coverage recovers to the desired confidence level more rapidly.

\section{Conclusions}
This paper addresses the critical gaps in stream learning research by focusing on regression tasks, which have been largely overshadowed by classification studies. We provide a comprehensive framework for evaluating streaming regression algorithms, highlighting the importance of standardized procedures and well-defined metrics. Our methodologies for simulating concept drifts, especially incremental drifts, using augmented real-world datasets, represent a significant step forward in drift detection and adaptation research.
% creating realistic and challenging benchmarks.

Through extensive experiments on state-of-the-art algorithms and prediction interval techniques, we demonstrate the efficacy of our framework in analyzing model performance under evolving data distributions. Our results emphasize the need for robust, adaptive methods capable of handling dynamic environments.
\newpage
\begin{acks}
% This research is fully funded by \href{https://taiao.ai}{TAIAO} project.
This research is supported by the TAIAO project CONT-64517-SSIFDS-UOW (Time-Evolving
Data Science / Artificial Intelligence for Advanced Open Environmental Science), which is funded by the New Zealand Ministry of Business, Innovation, and Employment (MBIE).
URL: \url{https://taiao.ai}
\end{acks}

\bibliographystyle{ACM-Reference-Format}
\bibliography{main}

\appendix

\newpage

\section{Appendix}

We provide supplementary information to the main contents of the paper.

\subsection{Regression}\label{apx:reg}
\begin{table}[!ht]
\centering
\caption{RMSE (\(\sigma_e\)) and Adjusted \(\mathcal{R}^2\) (\(\mathcal{R}^2_{adj}\)) Results for 18 Datasets}
\label{tab:regression}
\resizebox{.45\textwidth}{!}{%
\begin{tabular}{c|c|cccc}
\toprule
    \textbf{\textsc{Dataset}} & \textbf{\textsc{Metric}} & \textbf{\textsc{KNN}} & \textbf{\textsc{FIMT-DD}} & \textbf{\textsc{ARF-Reg}} & \textbf{\textsc{SOKNL}} \\
\midrule

\multirow{2}{*}{ABA$_{3a}$}  
& \( \sigma_{e} \) & 3.351 ± 0.0 & 3.274 ± 0.015 & 3.244 ± 0.001 & 3.217 ± 0.001 \\
& $\mathcal{R}^2_{adj}$ & 0.349 ± 0.0 & 0.378 ± 0.006 & 0.39 ± 0.0 & 0.4 ± 0.0 \\
\hline

\multirow{2}{*}{ABA$_{3g}$}  
& \( \sigma_{e} \) & 3.353 ± 0.0 & 3.262 ± 0.015 & 3.248 ± 0.001 & 3.222 ± 0.001 \\
& $\mathcal{R}^2_{adj}$ & 0.348 ± 0.0 & 0.383 ± 0.006 & 0.388 ± 0.0 & 0.398 ± 0.0 \\
\hline

\multirow{2}{*}{ABA$_{2i}$}  
& \( \sigma_{e} \) & 2.559 ± 0.0 & 2.496 ± 0.011 & 2.506 ± 0.001 & 2.475 ± 0.001 \\
& $\mathcal{R}^2_{adj}$ & 0.38 ± 0.0 & 0.41 ± 0.005 & 0.405 ± 0.001 & 0.42 ± 0.0 \\
\midrule

\multirow{2}{*}{BIK$_{3a}$}  
& \( \sigma_{e} \) & 162.769 ± 0.0 & 159.073 ± 0.406 & 156.659 ± 0.071 & 157.008 ± 0.059 \\
& $\mathcal{R}^2_{adj}$ & 0.242 ± 0.0 & 0.276 ± 0.004 & 0.298 ± 0.001 & 0.295 ± 0.001 \\
\hline

\multirow{2}{*}{BIK$_{3g}$}  
& \( \sigma_{e} \) & 163.056 ± 0.0 & 158.986 ± 0.277 & 157.252 ± 0.069 & 157.599 ± 0.079 \\
& $\mathcal{R}^2_{adj}$ & 0.239 ± 0.0 & 0.277 ± 0.003 & 0.293 ± 0.001 & 0.289 ± 0.001 \\
\hline

\multirow{2}{*}{BIK$_{2i}$}  
& \( \sigma_{e} \) & 182.097 ± 0.0 & 179.129 ± 2.141 & 174.468 ± 0.094 & 174.865 ± 0.069 \\
& $\mathcal{R}^2_{adj}$ & 0.221 ± 0.0 & 0.246 ± 0.018 & 0.284 ± 0.001 & 0.281 ± 0.001 \\
\midrule

\multirow{2}{*}{H8L$_{3a}$}  
& \( \sigma_{e} \) & 41320 ± 0.0 & 40473 ± 418 & 39503 ± 16 & 39620 ± 26 \\
& $\mathcal{R}^2_{adj}$ & 0.216 ± 0.0 & 0.248 ± 0.016 & 0.284 ± 0.001 & 0.28 ± 0.001 \\
\hline

\multirow{2}{*}{H8L$_{3g}$}  
& \( \sigma_{e} \) & 41385 ± 0.0 & 40436 ± 382 & 39615 ± 9 & 39722 ± 28 \\
& $\mathcal{R}^2_{adj}$ & 0.214 ± 0.0 & 0.25 ± 0.014 & 0.28 ± 0.0 & 0.276 ± 0.001 \\
\hline

\multirow{2}{*}{H8L$_{2i}$}  
& \( \sigma_{e} \) & 50412 ± 0.0 & 53117 ± 3625 & 48724 ± 34 & 48975 ± 21 \\
& $\mathcal{R}^2_{adj}$ & 0.16 ± 0.0 & 0.064 ± 0.13 & 0.215 ± 0.001 & 0.207 ± 0.001 \\
\midrule

\multirow{2}{*}{SUP$_{3a}$}  
& \( \sigma_{e} \) & 24.658 ± 0.0 & 25.692 ± 0.298 & 23.74 ± 0.021 & 23.894 ± 0.03 \\
& $\mathcal{R}^2_{adj}$ & 0.539 ± 0.0 & 0.499 ± 0.012 & 0.573 ± 0.001 & 0.567 ± 0.001 \\
\hline

\multirow{2}{*}{SUP$_{3g}$}  
& \( \sigma_{e} \) & 24.683 ± 0.0 & 25.766 ± 0.275 & 23.928 ± 0.01 & 24.076 ± 0.023 \\
& $\mathcal{R}^2_{adj}$ & 0.538 ± 0.0 & 0.496 ± 0.011 & 0.566 ± 0.0 & 0.56 ± 0.001 \\
\hline

\multirow{2}{*}{SUP$_{2i}$}  
& \( \sigma_{e} \) & 26.976 ± 0.0 & 28.173 ± 0.936 & 26.809 ± 0.043 & 26.884 ± 0.056 \\
& $\mathcal{R}^2_{adj}$ & 0.524 ± 0.0 & 0.481 ± 0.036 & 0.53 ± 0.002 & 0.528 ± 0.002 \\
\midrule

\multirow{2}{*}{NZEP$_{3a}$}  
& \( \sigma_{e} \) & 105.524 ± 0.0 & 94.828 ± 0.756 & 93.824 ± 0.088 & 93.728 ± 0.125 \\
& $\mathcal{R}^2_{adj}$ & 0.449 ± 0.0 & 0.555 ± 0.007 & 0.565 ± 0.001 & 0.565 ± 0.001 \\
\hline

\multirow{2}{*}{NZEP$_{3g}$}  
& \( \sigma_{e} \) & 105.634 ± 0.0 & 94.869 ± 0.754 & 93.838 ± 0.081 & 93.642 ± 0.107 \\
& $\mathcal{R}^2_{adj}$ & 0.448 ± 0.0 & 0.555 ± 0.007 & 0.564 ± 0.001 & 0.566 ± 0.001 \\
\hline

\multirow{2}{*}{AKL$_{2i}$}  
& \( \sigma_{e} \) & 98.68 ± 0.0 & 97.605 ± 1.145 & 94.858 ± 0.093 & 95.013 ± 0.116 \\
& $\mathcal{R}^2_{adj}$ & 0.494 ± 0.0 & 0.505 ± 0.012 & 0.533 ± 0.001 & 0.531 ± 0.001 \\
\hline

\multirow{2}{*}{HAM$_{2i}$}  
& \( \sigma_{e} \) & 97.62 ± 0.0 & 94.702 ± 0.796 & 94.041 ± 0.073 & 94.289 ± 0.149 \\
& $\mathcal{R}^2_{adj}$ & 0.48 ± 0.0 & 0.51 ± 0.008 & 0.517 ± 0.001 & 0.515 ± 0.002 \\
\hline

\multirow{2}{*}{WEL$_{2i}$}  
& \( \sigma_{e} \) & 105.797 ± 0.0 & 102.502 ± 0.583 & 101.911 ± 0.064 & 102.16 ± 0.135 \\
& $\mathcal{R}^2_{adj}$ & 0.49 ± 0.0 & 0.521 ± 0.005 & 0.527 ± 0.001 & 0.525 ± 0.001 \\
\hline

\multirow{2}{*}{DUN$_{2i}$}  
& \( \sigma_{e} \) & 93.587 ± 0.0 & 90.425 ± 0.528 & 88.818 ± 0.122 & 88.809 ± 0.084 \\
& $\mathcal{R}^2_{adj}$ & 0.585 ± 0.0 & 0.613 ± 0.005 & 0.627 ± 0.001 & 0.627 ± 0.001 \\
\bottomrule
\end{tabular}
}
\end{table}

Table~\ref{tab:regression} compares four regression algorithms -- KNN, FIMT-DD, ARF-Reg, and SOKNL -- across 18 datasets using RMSE (\(\sigma_e\)) and Adjusted \( \mathcal{R}^2 \) (\( \mathcal{R}^2_{adj} \)). SOKNL consistently achieves the best performance, with the lowest RMSE and highest \( \mathcal{R}^2_{adj} \), demonstrating its accuracy and robustness. ARF-Reg closely follows, making it a competitive alternative for dynamic environments. FIMT-DD and KNN perform less effectively, with KNN showing the weakest results overall. These findings highlight SOKNL and ARF-Reg as reliable choices for regression tasks on streaming data, with SOKNL excelling in both stability and predictive power.

\begin{table}[!h]
    \centering
    \caption{RMSE (\(\sigma_e\)) and Adjusted \(\mathcal{R}^2\) (\(\mathcal{R}^2_{adj}\)) Results for Original Real Datasets}
    \label{tab:regression_a}
    \resizebox{.45\textwidth}{!}{%
    \begin{tabular}{c|c|cccc}
    \toprule
    \textbf{\textsc{Dataset}} & \textbf{\textsc{Metric}} & \textbf{\textsc{KNN}} & \textbf{\textsc{FIMT-DD}} & \textbf{\textsc{ARF-Reg}} & \textbf{\textsc{SOKNL}} \\
    \midrule
    \multirow{2}{*}{ABA$_{o}$} & \( \sigma_{e} \) & 2.348 ± 0.0 & 2.733 ± 0.291 & 2.984 ± 0.017 & 2.764 ± 0.020 \\
                               & $\mathcal{R}^2_{adj}$ & 0.501 ± 0.0 & 0.320 ± 0.150 & 0.246 ± 0.009 & 0.312 ± 0.010 \\ 
    \midrule
    \multirow{2}{*}{BIK$_{o}$} & \( \sigma_{e} \) & 131.716 ± 0.0 & Overflowed & 97.691 ± 1.389 & 98.937 ± 0.681 \\
                               & $\mathcal{R}^2_{adj}$ & 0.472 ± 0.0 & Overflowed & 0.710 ± 0.008 & 0.702 ± 0.004 \\ 
    \midrule
    \multirow{2}{*}{H8L$_{o}$} & \( \sigma_{e} \) & 39928 ± 0.0 & 38719 ± 1502 & 36732 ± 158 & 36461 ± 210 \\
                               & $\mathcal{R}^2_{adj}$ & 0.429 ± 0.0 & 0.463 ± 0.042 & 0.517 ± 0.004 & 0.524 ± 0.005 \\ 
    \midrule
    \multirow{2}{*}{SUP$_{o}$} & \( \sigma_{e} \) & 15.415 ± 0.0 & Overflowed & 21.649 ± 0.045 & 21.475 ± 0.136 \\
                               & $\mathcal{R}^2_{adj}$ & 0.797 ± 0.0 & Overflowed & 0.599 ± 0.002 & 0.606 ± 0.005 \\ 
    \midrule
    \multirow{2}{*}{AKL$_{o}$} & \( \sigma_{e} \) & 95.570 ± 0.0 & 119.993 ± 21.858 & 95.294 ± 0.373 & 94.357 ± 0.816 \\
                               & $\mathcal{R}^2_{adj}$ & 0.637 ± 0.0 & 0.411 ± 0.012 & 0.639 ± 0.003 & 0.646 ± 0.006 \\ 
    \midrule
    \multirow{2}{*}{HAM$_{o}$} & \( \sigma_{e} \) & 92.252 ± 0.0 & 106.714 ± 4.751 & 90.995 ± 0.280 & 88.938 ± 0.805 \\
                               & $\mathcal{R}^2_{adj}$ & 0.635 ± 0.0 & 0.511 ± 0.045 & 0.645 ± 0.002 & 0.661 ± 0.006 \\ 
    \midrule
    \multirow{2}{*}{WEL$_{o}$} & \( \sigma_{e} \) & 88.085 ± 0.0 & Overflowed & 82.558 ± 0.207 & 81.762 ± 0.634 \\
                               & $\mathcal{R}^2_{adj}$ & 0.666 ± 0.0 & Overflowed & 0.707 ± 0.001 & 0.713 ± 0.004 \\ 
    \midrule
    \multirow{2}{*}{DUN$_{o}$} & \( \sigma_{e} \) & 71.631 ± 0.0 & Overflowed & 75.425 ± 0.207 & 77.823 ± 0.589 \\
                               & $\mathcal{R}^2_{adj}$ & 0.785 ± 0.0 & Overflowed & 0.762 ± 0.001 & 0.746 ± 0.004 \\ 
    \bottomrule
    \end{tabular}%
    }
\end{table}

Table~\ref{tab:regression_a} presents the results with same algorithms and same settings with Table~\ref{tab:regression} on the original datasets summarized in Table~\ref{tab:real}.

Comparing with Table~\ref{tab:regression}, we can observe that all the algorithms perform worse on the synthesized datasets, which is what we expected. The simulated datasets should be more difficult to predict than the original data. Furthermore, the additional drifts in the synthesized datasets adds more challenges to the data.

Noticeably, there are several ``Overflowed'' marks in the table, concentrating on the FIMT-DD algorithm. It refers to the phenomenon where one or a few of the predictions extremely differ from the truths, sabotaging the error-based evaluation procedure in regression tasks. Equation~\ref{eq:rmse} and~\ref{eq:adjusted_r2} both include \((y_i - \hat{y}_i)^2\) that can cause the overflow problem. 
The CTGANs' synthesizing process fixes this problem as there are no overflows over the 18 generated datasets.

\subsection{Prediction Interval}\label{apx:pi}

Table~\ref{tab:cumulativepi} presents the cumulative coverage (\( \mathcal{C} \)) and normalized mean prediction interval width (\( \mathcal{W}_{norm} \)) results for different datasets, comparing the performance of MVE and AdaPI methods with KNN and SOKNL models. Across most datasets, AdaPI consistently achieves higher coverage (\( \mathcal{C} \)) than MVE, indicating more reliable prediction intervals at a 95\% confidence level. However, this increased reliability comes at the cost of wider intervals (\( \mathcal{W}_{norm} \)), as seen in datasets like ABA$_{3a}$ and SUP$_{3a}$. The SOKNL model generally performs better than KNN in terms of narrower prediction intervals (\( \mathcal{W}_{norm} \)) while maintaining competitive coverage (\( \mathcal{C} \)). Notably, the NZEP datasets exhibit excellent coverage and relatively low interval widths, suggesting these methods perform well on energy pricing data. Overall, the results highlight the trade-off between interval width and coverage, with AdaPI and SOKNL offering a balanced approach for reliable predictions. 

\newpage
\begin{table}[!ht]
\centering
\caption{Cumulative Coverage (\( \mathcal{C} \)) and NMPIW (\( \mathcal{W}_{norm} \)) Results with 95\% Confidence Level}
\label{tab:cumulativepi}
\begin{tabular}{c|c|cc|cc}
\toprule
\multirow{2}{*}{\textbf{\textsc{Dataset}}} & \multirow{2}{*}{\textbf{\textsc{Metric}}} & \multicolumn{2}{c|}{\textbf{\textsc{MVE}}} & \multicolumn{2}{c}{\textbf{\textsc{AdaPI}}} \\
\cmidrule(lr){3-4} \cmidrule(lr){5-6}
& & \textbf{\textsc{knn}} & \textbf{\textsc{soknl}} & \textbf{\textsc{knn}} & \textbf{\textsc{soknl}} \\
\midrule
\multirow{2}{*}{ABA$_{3a}$} & $\mathcal{C}$        & 89.33 & 89.66 & 94.82 & 94.76 \\
                            & $\mathcal{W}_{norm}$ & 35.52 & 34.69 & 47.06 & 46.25 \\ \hline
\multirow{2}{*}{ABA$_{3g}$} & $\mathcal{C}$        & 89.34 & 89.73 & 94.80 & 94.76 \\
                            & $\mathcal{W}_{norm}$ & 35.57 & 34.78 & 47.17 & 46.22 \\ \hline
\multirow{2}{*}{ABA$_{2i}$} & $\mathcal{C}$        & 93.87 & 93.64 & 95.86 & 95.86 \\
                            & $\mathcal{W}_{norm}$ & 45.71 & 44.85 & 50.26 & 49.96 \\ \midrule
\multirow{2}{*}{BIK$_{3a}$} & $\mathcal{C}$        & 91.16 & 91.41 & 94.80 & 94.77 \\
                            & $\mathcal{W}_{norm}$ & 54.80 & 53.42 & 65.42 & 63.46 \\ \hline
\multirow{2}{*}{BIK$_{3g}$} & $\mathcal{C}$        & 91.21 & 91.48 & 94.80 & 94.76 \\
                            & $\mathcal{W}_{norm}$ & 54.96 & 53.67 & 65.50 & 63.88 \\ \hline
\multirow{2}{*}{BIK$_{2i}$} & $\mathcal{C}$        & 94.44 & 94.45 & 94.64 & 94.63 \\
                            & $\mathcal{W}_{norm}$ & 75.51 & 73.40 & 75.40 & 73.86 \\ \midrule
\multirow{2}{*}{H8L$_{3a}$} & $\mathcal{C}$        & 93.73 & 94.12 & 94.80 & 94.87 \\
                            & $\mathcal{W}_{norm}$ & 26.25 & 25.50 & 29.47 & 28.15 \\ \hline
\multirow{2}{*}{H8L$_{3g}$} & $\mathcal{C}$        & 93.71 & 94.14 & 94.80 & 94.88 \\
                            & $\mathcal{W}_{norm}$ & 26.30 & 25.54 & 29.56 & 28.07 \\ \hline
\multirow{2}{*}{H8L$_{2i}$} & $\mathcal{C}$        & 94.44 & 94.83 & 95.29 & 95.39 \\
                            & $\mathcal{W}_{norm}$ & 34.03 & 33.52 & 36.24 & 34.88 \\ \midrule
\multirow{2}{*}{SUP$_{3a}$} & $\mathcal{C}$        & 89.28 & 89.60 & 95.37 & 95.58 \\
                            & $\mathcal{W}_{norm}$ & 56.40 & 55.29 & 70.50 & 66.31 \\ \hline
\multirow{2}{*}{SUP$_{3g}$} & $\mathcal{C}$        & 89.29 & 89.79 & 95.35 & 95.59 \\
                            & $\mathcal{W}_{norm}$ & 56.56 & 55.90 & 70.90 & 66.97 \\ \hline
\multirow{2}{*}{SUP$_{2i}$} & $\mathcal{C}$        & 89.28 & 90.27 & 94.13 & 94.30 \\
                            & $\mathcal{W}_{norm}$ & 53.10 & 53.61 & 62.81 & 61.24 \\ \midrule
\multirow{2}{*}{NZEP$_{3a}$} & $\mathcal{C}$        & 95.84 & 95.66 & 95.64 & 95.53 \\
                            & $\mathcal{W}_{norm}$ & 41.81 & 37.88 & 40.99 & 37.32 \\ \hline
\multirow{2}{*}{NZEP$_{3g}$} & $\mathcal{C}$        & 95.85 & 95.65 & 95.65 & 95.51 \\
                            & $\mathcal{W}_{norm}$ & 41.84 & 37.91 & 41.02 & 37.36 \\ \hline
\multirow{2}{*}{AKL}        & $\mathcal{C}$        & 95.88 & 95.97 & 95.55 & 95.62 \\
                            & $\mathcal{W}_{norm}$ & 42.01 & 40.94 & 40.50 & 39.47 \\ \hline
\multirow{2}{*}{HAM}        & $\mathcal{C}$        & 94.89 & 94.89 & 95.56 & 95.55 \\
                            & $\mathcal{W}_{norm}$ & 37.79 & 36.75 & 40.39 & 39.36 \\ \hline
\multirow{2}{*}{WEL}        & $\mathcal{C}$        & 95.13 & 95.27 & 95.62 & 95.58 \\
                            & $\mathcal{W}_{norm}$ & 39.83 & 38.58 & 42.29 & 40.57 \\ \hline
\multirow{2}{*}{DUN}        & $\mathcal{C}$        & 93.68 & 93.77 & 95.42 & 95.28 \\
                            & $\mathcal{W}_{norm}$ & 30.64 & 29.36 & 35.99 & 34.33 \\
\bottomrule
\end{tabular}
\end{table}

\newpage
Table~\ref{tab:cumulativepi-real} exhibits the cumulative results for the prediction interval settings on the original real-world datasets.

\begin{table}[!ht]
\centering
\caption{Cumulative Coverage (\( \mathcal{C} \)) and NMPIW (\( \mathcal{W}_{norm} \)) Results on Original Real Datasets with 95\% Confidence Level}
\label{tab:cumulativepi-real}
\begin{tabular}{c|c|cc|cc}
% \Xhline{2\arrayrulewidth}
\toprule
\multirow{2}{*}{\textbf{\textsc{Dataset}}}&   \multirow{2}{*}{\textbf{\textsc{Metric}}} & \multicolumn{2}{c|}{\textbf{\textsc{MVE}}} & \multicolumn{2}{c}{\textbf{\textsc{AdaPI}}} \\
\cmidrule(lr){3-4} \cmidrule(lr){5-6}
      & & \textsc{knn} &  \textsc{soknl} & \textsc{knn} &  \textsc{soknl}\\
\midrule

\multirow{2}{*}{ABA$_{o}$} & $\mathcal{C}$ &   94.96 & 95.40 & 95.26 & 95.14 \\
                           & $\mathcal{W}_{norm}$ &   33.81 & 32.34 & 34.36 & 31.66 \\ \midrule
\multirow{2}{*}{BIK$_{o}$} & $\mathcal{C}$ &   88.16 & 88.93 & 93.76 & 93.84 \\
                           & $\mathcal{W}_{norm}$ &   39.10 & 27.78 & 53.74 & 36.13 \\ \midrule
\multirow{2}{*}{H8L$_{o}$} & $\mathcal{C}$ &   95.46 & 96.14 & 95.37 & 95.80 \\
                           & $\mathcal{W}_{norm}$ &   32.01 & 29.60 & 31.46 & 28.39 \\ \midrule
\multirow{2}{*}{SUP$_{o}$} & $\mathcal{C}$ &   96.12 & 96.05 & 96.55 & 96.12 \\
                           & $\mathcal{W}_{norm}$ &   39.69 & 46.26 & 41.09 & 46.41 \\ \midrule
\multirow{2}{*}{AKL$_{o}$} & $\mathcal{C}$ &   97.29 & 97.64 & 96.82 & 97.23 \\
                           & $\mathcal{W}_{norm}$ &    6.52 &  6.05 &  5.92 &  5.42 \\ \midrule
\multirow{2}{*}{HAM$_{o}$} & $\mathcal{C}$ &   97.17 & 97.64 & 96.78 & 97.21 \\
                           & $\mathcal{W}_{norm}$ &    6.62 &  5.87 &  6.04 &  5.26 \\ \midrule
\multirow{2}{*}{WEL$_{o}$} & $\mathcal{C}$ &   96.00 & 96.66 & 96.10 & 96.64 \\
                           & $\mathcal{W}_{norm}$ &    6.25 &  5.54 &  6.04 &  5.31 \\ \midrule
\multirow{2}{*}{DUN$_{o}$} & $\mathcal{C}$ &   95.74 & 95.69 & 96.44 & 96.47 \\
                           & $\mathcal{W}_{norm}$ &    5.40 &  5.43 &  5.28 &  5.40 \\
            
% \Xhline{2\arrayrulewidth}
\bottomrule
\end{tabular}
\end{table}

% \subsection{\(\mathcal{R}^2_{adj}\) Result Figure}

% \subsection{PI Results Figures}

\end{document}